\documentclass[final,3p,times,twocolumn]{elsarticle}

\usepackage{graphicx}
\usepackage{amsmath}
\usepackage{amsfonts}
\usepackage[colorlinks=True]{hyperref}
\usepackage[all]{hypcap}
\usepackage{comment}
\usepackage{lineno}
\modulolinenumbers[1]
\usepackage{booktabs}
\usepackage{multirow}
\usepackage{float}
\usepackage{color, soul}

\journal{arXiv}

\biboptions{numbers,sort&compress}

\begin{document}

\makeatletter
\def\ps@pprintTitle{%
  \let\@oddhead\@empty
  \let\@evenhead\@empty
  \let\@oddfoot\@empty
  \let\@evenfoot\@oddfoot
}
\makeatother

\begin{frontmatter}

\title{Robust Table Detection and Structure Recognition from \\ Heterogeneous Document Images}

\author[address1]{Chixiang Ma\fnref{myfootnote}\corref{chixma}}
\ead{chixiangma@gmail.com}

\author[address2]{Weihong Lin}
\ead{weihlin@microsoft.com}

\author[address2]{Lei Sun}
\ead{lsun@microsoft.com}

\author[address2]{Qiang Huo}
\ead{qianghuo@microsoft.com}

\address[address1]{Department of EEIS, University of Science and Technology of China, Hefei, 230026, China}
\address[address2]{Microsoft Research Asia, Beijing, 100080, China}
\fntext[myfootnote]{This work was done when Chixiang Ma was an intern in MMI Group, Microsoft Research Asia, Beijing, China.}
\cortext[chixma]{Corresponding author.}

\begin{abstract}
We introduce a new table detection and structure recognition approach named RobusTabNet to detect the boundaries of tables and reconstruct the cellular structure of each table from heterogeneous document images. For table detection, we propose to use CornerNet as a new region proposal network to generate higher quality table proposals for Faster R-CNN, which has significantly improved the localization accuracy of Faster R-CNN for table detection. Consequently, our table detection approach achieves state-of-the-art performance on three public table detection benchmarks, namely cTDaR TrackA, PubLayNet and IIIT-AR-13K, by only using a lightweight ResNet-18 backbone network. Furthermore, we propose a new split-and-merge based table structure recognition approach, in which a novel spatial CNN based separation line prediction module is proposed to split each detected table into a grid of cells, and a Grid CNN based cell merging module is applied to recover the spanning cells. As the spatial CNN module can effectively propagate contextual information across the whole table image, our table structure recognizer can robustly recognize tables with large blank spaces and geometrically distorted (even curved) tables. Thanks to these two techniques, our table structure recognition approach achieves state-of-the-art performance on three public benchmarks, including SciTSR, PubTabNet and {\color{black}{cTDaR TrackB2-Modern}}. Moreover, we have further demonstrated the advantages of our approach in recognizing tables with complex structures, large blank spaces, as well as geometrically distorted or even curved shapes on a more challenging in-house dataset.
\end{abstract}

\begin{keyword}
Table detection \sep Table structure recognition \sep Corner detection \sep Spatial CNN \sep Grid CNN \sep Split-and-merge
\end{keyword}

\end{frontmatter}


\section{Introduction}
\label{sec:intro}
Tables are a prevalent means of representing and communicating structured data, which are widely used in diverse types of documents including financial statements, scientific papers, invoices, purchasing orders, etc. With the explosive growth of the number of documents, automatic table detection and structure recognition techniques are eagerly desired to reconstruct tables from document images, which can facilitate many downstream applications, such as information retrieval \cite{liu2007tableseer} and question answering \cite{sun2016table}. The aim of table detection is to detect the boundaries of tables, while the aim of table structure recognition (TSR) is to reconstruct the cellular structure of each detected table, i.e., identifying the coordinates of each cell bounding box as well as its row and column spanning information. Both table detection and structure recognition are unsolved problems due to the following challenges.
{\color{black}{First, tables in documents may have complex structures and diverse styles (erratic use of ruling lines). For example, in financial reports, some borderless tables may have complex hierarchical header structures, contain many empty or spanning cells, or have extremely large/small blank spaces between neighboring columns. Some neighboring tables may be very close to each other, making it hard to determine whether they should be merged or not. In invoices, tables may have different sizes, e.g., some line-item tables may only contain two rows and some others may span multiple pages. Second, tables cells may contain diverse contents, ranging from a single character to a set of more complex page objects such as paragraphs, tables, figures, formulas, etc.
Third, some background objects in documents like figures, graphics, flow charts and structurally laid out texts, may have similar textures as tables, which poses another challenge for reduction of false alarms. In forms, some tables may be embedded in other more complex tabular objects (e.g., nested tables), which makes table boundaries ambiguous.
Moreover, many camera-captured document images are of poor image quality, and tables contained in them may be distorted (even curved) or contain artifacts or noises, which makes table detection and structure recognition even more difficult. }}

In recent years, the success of deep learning in various computer vision applications has motivated researchers to explore deep neural networks for detecting tables and recognizing table structures from document images. These deep learning based table detection and structure recognition approaches have substantially outperformed traditional rule or statistical machine learning based methods in terms of both accuracy and capability \cite{hashmi2021current}. Most deep learning based table detection approaches (e.g., \cite{hao2016table,vo2018ensemble,gilani2017table,huang2019yolo,zheng2021global,saha2019graphical,prasad2020cascadetabnet,agarwal2021cdec}) treat table as a specific object and borrow various CNN-based object detection and segmentation frameworks, like Faster R-CNN \cite{ren2015faster}, Mask R-CNN \cite{he2017mask}, and Cascade R-CNN \cite{cai2019cascade}, to solve the table detection problem. With the help of some effective techniques like more powerful backbone networks and deformable convolution operations, these CNN based table detection methods, especially CDeC-Net \cite{agarwal2021cdec}, have achieved superior performance on many public table detection benchmark datasets. {\color{black}{Despite this, the localization accuracy of these methods is still far from satisfactory. For instance, although CDeC-Net has leveraged the Cascade R-CNN model \cite{cai2019cascade} to improve table detection accuracy, its detection accuracy still drops a lot when the Intersection-over-Union (IoU) threshold is increased from 0.5 to 0.9 during evaluation (Table VIII in \cite{agarwal2021cdec}). As the localization accuracy of table detection will significantly affect the performance of the following TSR task, more effective techniques to improve the localization accuracy of these CNN based table detection methods are still desired.}} For table structure recognition, deep learning based methods (e.g., \cite{zheng2021global,schreiber2017deepdesrt,siddiqui2019rethinking,tensmeyer2019deep,khan2019table,siddiqui2019deeptabstr,hashmi2021guided,qasim2019rethinking,raja2020table,qiao2021lgpma,li2021adaptive}) have already made great progress towards recognizing tables with complex structures and diverse styles. Recent best performing table structure recognition approaches, like TabStruct-Net \cite{raja2020table} and LGPMA \cite{qiao2021lgpma}, typically use CNN-based object detection or segmentation models like Mask R-CNN to detect table cells first, then adopt some cell grouping/clustering algorithms to predict row/column relationships between the detected cells. {\color{black}{Although these methods have achieved very high accuracy on benchmark datasets like SciTSR \cite{chi2019complicated} and PubTabNet \cite{zhong2020image}, they still cannot be directly applied to geometrically distorted or even curved tables as they rely on an assumption that tables are axis-aligned.}} In some real-world application scenarios like the “Insert data from picture” feature\footnote{\url{https://support.microsoft.com/en-us/office/insert-data-from-picture-3c1bb58d-2c59-4bc0-b04a-\\a671a6868fd7}} in Excel, document images may be captured by mobile cameras. In these camera-captured images, it is inevitable that tables are geometrically distorted. However, existing benchmark datasets haven’t taken this important scenario into account as images in these datasets are either captured by scanners or converted from digital PDF files. Thus, more research is needed to find out new table structure recognition approaches robust to geometrically distorted or even curved tables.

In this paper, we propose a new table detection and structure recognition approach named RobusTabNet to overcome the abovementioned challenges. For table detection, we use CornerNet \cite{law2018cornernet} as a new region proposal network for Faster R-CNN, which generates table proposals by detecting and grouping corner points. With these corner-based high quality region proposals, our approach achieves superior performance even with a very lightweight backbone network, i.e., ResNet-18 \cite{he2016deep}. For TSR, we present a new split-and-merge based approach and propose two effective techniques to significantly improve its capability. First, we propose a novel spatial CNN \cite{pan2018spatial} based separation line prediction module to split each detected table into a grid of cells. As the spatial CNN can effectively propagate contextual information across the whole table image, our separation line prediction algorithm can improve the robustness of our table structure recognizer to tables with large blank spaces and distorted or even curved shapes. Second, we propose a Grid CNN based cell merging module to recover the wrongly split cells, especially the spanning cells. In this module, the whole table is compactly represented as a grid so that a simple CNN based cell merging module can achieve higher accuracy than Relation Network or Graph Convolutional Network (GCN) based methods. With these new techniques, the proposed RobusTabNet has achieved state-of-the-art performance on both table detection (cTDaR TrackA \cite{gao2019icdar}, PubLayNet \cite{zhong2019publaynet} and IIIT-AR-13K \cite{mondal2020iiit}) and structure recognition (SciTSR \cite{chi2019complicated}, PubTabNet \cite{zhong2020image} and {\color{black}{cTDaR TrackB2-Modern}} \cite{gao2019icdar}) public benchmarks. We have further validated the robustness of our approach to tables with complex structures, large blank spaces, as well as distorted or even curved shapes on a more challenging in-house dataset.

{\color{black}{The main contributions of this paper are as follows:
\begin{itemize}
\item We present a new table detector by using CornerNet as a new region proposal network for Faster R-CNN to achieve high table localization accuracy. Compared with RPN \cite{ren2015faster}, the percentage of well-localized proposals (IoU$>$0.9) in the positive samples (IoU$>$0.7) from CornerNet is much higher, which contributes to better end-to-end table detection performance.

\item We present a new split-and-merge based table structure recognizer, which is robust to geometrically distorted or even curved tables. To this end, a new spatial CNN based separation line prediction approach is proposed to robustly predict curvilinear separation lines from distorted or even curved tables, while a Grid CNN module is proposed to recover spanning cells efficiently and effectively.

\item Our proposed table extraction approach, RobusTabNet, has achieved state-of-the-art performance on both table detection (cTDaR TrackA, PubLayNet and IIIT-AR-13K) and structure recognition (SciTSR, PubTabNet and {\color{black}{cTDaR TrackB2-Modern}}) public benchmarks.

\end{itemize}
}}

\section{Related work}
\label{sec:related_work}
\subsection{Table detection}
\label{subsec:TD}
\subsubsection{Traditional methods}
Rule-based methods are among the earliest approaches for locating tables inside documents. These methods usually exploit visual clues (e.g., text-block arrangement \cite{kieninger1998t}, or horizontal and vertical lines \cite{gatos2005automatic,hassan2007table,anh2015hybrid}), keywords \cite{tupaj1996extracting,harit2012table}, or formal templates \cite{wangt2001automatic} to detect tables in particular scenarios. We refer readers to \cite{zanibbi2004survey,embley2006table} for a more detailed summarization of these conventional approaches. Rule-based methods usually require extensive manual efforts to design heuristic rules and tune hyper-parameters. To reduce the dependence on heuristics, lots of statistical machine learning based approaches have been proposed, e.g., \cite{cesarini2002trainable,e2009learning}.
Although these methods have improved table detection accuracy significantly, they still rely on handcrafted features, which limit their generalization ability. A comprehensive review of these statistical machine learning based methods can be found in \cite{jorge2006design}.

\subsubsection{Deep learning based methods}
With the rapid development of deep learning, numerous CNN based table detection methods have been proposed and outperformed traditional methods by a big margin in terms of both accuracy and capability. These methods can be roughly classified into three categories: object detection based methods, semantic segmentation based methods, and bottom-up methods.

\textbf{Object detection based methods.} These methods adapt state-of-the-art top-down object detection or instance segmentation frameworks to solve the table detection problem. Initially, Hao et al. \cite{hao2016table}, Yi et al. \cite{yi2017cnn}, and Oliveira et al. \cite{augusto2017fast} adopted R-CNN \cite{girshick2014rich} for table detection first, but the performance of these methods was limited by the traditional region proposal generation methods, which relied on the heuristic rules and handcrafted features. Later, more advanced object detectors, like Fast R-CNN \cite{girshick2015fast}, Faster R-CNN \cite{ren2015faster}, YOLO \cite{redmon2016you}, RetinaNet \cite{lin2017focal}, Mask R-CNN \cite{he2017mask}, Cascade Mask R-CNN \cite{cai2019cascade}, were explored by Vo et al. \cite{vo2018ensemble}, Gilani et al. \cite{gilani2017table}, Schreiber et al. \cite{schreiber2017deepdesrt}, Huang et al. \cite{huang2019yolo}, Zheng et al. \cite{zheng2021global}, Saha et al. \cite{saha2019graphical}, Prasad et al. \cite{prasad2020cascadetabnet} and Agarwal et al. \cite{agarwal2021cdec}, to detect tables (as well as other page objects like figures and formulas) from document images, respectively. The accuracy of these detectors for table detection could be improved further by adding some effective techniques. For example, Gilani et al. \cite{gilani2017table}, Arif et al. \cite{arif2018table}, and Prasad et al. \cite{prasad2020cascadetabnet} proposed to use image transformation techniques, e.g., distance transforms, coloration and dilation, to enhance input document images or augment the used training sets so that additional clues could be provided to the detectors. Siddiqui et al. \cite{siddiqui2018decnt} incorporated deformable convolution and deformable RoI Pooling operations \cite{dai2017deformable} into Faster R-CNN to make the model more robust to geometric transformations. Agarwal et al. \cite{agarwal2021cdec} employed a more powerful backbone network, i.e., a composite backbone network \cite{liu2020cbnet} with deformable convolution filters, to push the accuracy of Cascade Mask R-CNN further. {\color{black}{Although this method achieved state-of-the-art performance on several benchmark datasets (e.g., \cite{gobel2013icdar,gao2017icdar2017,gao2019icdar,zhong2019publaynet,mondal2020iiit,li2020tablebank})}}, it suffered from high computation complexity and memory usage. To improve the localization accuracy, Sun et al. \cite{sun2019faster} adopted Faster R-CNN to detect table boxes and the corresponding corner boxes simultaneously and used a post-processing algorithm to adjust table boundaries according to the detected corners. However, the corner boxes are manually predefined small boxes, and the size has no explicit meaning, which leads to higher miss detection rate for corners \cite{sun2019faster}. 

\textbf{Semantic segmentation based methods.} These methods (e.g., \cite{yang2017learning,he2017multi,kavasidis2019saliency,paliwal2019tablenet}) treat table detection as a semantic segmentation problem and leverage existing semantic segmentation frameworks like FCN \cite{long2015fully} to predict a pixel-level segmentation mask first, and then group table pixels into tables. Yang et al. \cite{yang2017learning} proposed a multimodal FCN for page segmentation to detect tables and other page objects, in which both visual features from images and linguistic features from the content of underlying texts are leveraged to improve segmentation accuracy. He et al. \cite{he2017multi} proposed a multi-scale multi-task FCN to predict two sets of segmentation masks for text-block/table/figure regions and their corresponding contours first. After refined by a conditional random field (CRF) model, these segmentation masks are then input to a post-processing module to obtain table regions. Kavasidis et al. \cite{kavasidis2019saliency} proposed a saliency-based FCN performing multi-scale reasoning on visual cues followed by a fully connected CRF for localizing tables and charts in digital/digitized documents. 

\textbf{Bottom-up methods.} Most bottom-up methods model each document image as a graph, where each node represents a page object (e.g., word, text-line) and each edge represents a neighboring relationship between two page objects, and then formulate table detection as a graph labeling problem. Li et al. \cite{li2018page} used traditional layout analysis methods to generate line regions first, then applied two hybrid CNN-CRF models to classify them into four classes (text, formula, table, figure) and predict whether each pair of line regions belong to a same cluster, respectively. After that, regions belonging to the same class and the same cluster were merged to get page objects. Riba et al. \cite{riba2022table} and Holeček et al. \cite{holevcek2019table} took text regions (words or text-lines) as nodes and generated a visibility or neighborhood graph to represent the underlying structure of each input document first, then used graph neural networks to perform node and edge classification. After that, connected subgraphs where the nodes are classified as table were extracted as tables. Recently, Li et al. \cite{li2020docbank} proposed to consider document layout analysis as a text-based sequence labeling problem and leveraged pre-trained language models to classify each word into a pre-defined page object category, including table. These bottom-up methods depend on certain assumptions like availability of accurate word/text-line bounding boxes as additional inputs.

\subsection{Table structure recognition}
\label{subsec:TSR}
\subsubsection{Traditional methods}
Early table structure recognition methods were mainly based on handcrafted features and heuristic rules. These methods (e.g., \cite{kieninger1998t,laurentini1992identifying,itonori1993table,shigarov2016configurable,rastan2019texus}) are mostly applied to simple table structures or specific data formats, such as PDF files. To reduce the dependence on heuristics, a few statistical machine learning based methods were proposed later, e.g., \cite{wang2004table}.
A comprehensive review of these traditional methods can be found in \cite{jorge2006design}. These traditional methods usually make strong assumptions about table layouts and rely on domain-specific heuristics, which limit their generalization ability.

\subsubsection{Deep learning based methods}
Recently, there is a trend to leverage deep learning models to solve the TSR problem. These methods can be roughly divided into three categories: row/column extraction based methods, image-to-markup generation based methods and bottom-up methods. 

\textbf{Row/column extraction based methods.} These methods usually adopt object detection or semantic segmentation frameworks to detect rows and columns from a table image first, then intersect the detected rows and columns to generate a grid of cells. DeepDeSRT \cite{schreiber2017deepdesrt} is the first to apply FCN based semantic segmentation models to the TSR task. They adopted two FCN models to segment tables into rows and columns first, and then used post-processing algorithms to deal with spurious detection fragments as well as severed and conjoined structures. However, this vanilla FCN based row/column segmentation method cannot robustly predict complete segmentation masks for rows and columns when tables contain large blank spaces \cite{siddiqui2019rethinking,tensmeyer2019deep}. To alleviate this problem, Siddiqui et al. \cite{siddiqui2019rethinking} and Tensmeyer et al. \cite{tensmeyer2019deep} pooled features along rows and columns of pixels on some intermediate feature maps, which enabled their FCN models to leverage much wider contextual information to improve row/column segmentation accuracy. Instead of relying on FCN, Khan et al. \cite{khan2019table} proposed to use sequential models like bi-directional gated recurrent unit networks (GRU) to scan pre-processed table images from top-to-bottom and left-to-right to identify row and column separators. Siddiqui et al. \cite{siddiqui2019deeptabstr} proposed to formulate the problem of row/column identification in a tabular structure as an object detection problem instead of a semantic segmentation problem, and leveraged three object detection models, namely deformable Faster R-CNN, deformable R-FCN and deformable FPN, to detect the bounding boxes of rows and columns from tables directly. Hashmi et al. \cite{hashmi2021guided} adopted another object detection model, i.e., Mask R-CNN with optimized anchors, to further improve row/column detection accuracy. All the abovementioned methods, except Tensmeyer et al. \cite{tensmeyer2019deep}, didn’t take spanning cells into consideration and can only recover the basic grid structures of tables. To deal with spanning cells, Tensmeyer et al. \cite{tensmeyer2019deep} presented the SPLERGE method, which used a Split model to produce the grid structure of an input table first, and then used a Merge model to predict which grid elements should be merged to recover spanning cells. Differing from this two-stage paradigm, Zou et al. \cite{zou2020deep} proposed a one-stage approach to segmenting the real row and column separators directly to avoid over-splitting spanning cells. {\color{black}{Although these methods have achieved promising results on some benchmark datasets, e.g., \cite{gobel2013icdar,gao2019icdar,siddiqui2019rethinking}}}, they cannot be directly applied to distorted or even curved tables as they rely on an implicit assumption that tables are axis-aligned.

\textbf{Image-to-markup generation based methods.} These methods treat table recognition as an image-to-markup generation problem and adopt existing image-to-markup models to directly convert each source table image into target presentational markup that fully describes its structure and cell contents. Deng et al. \cite{deng2019challenges} constructed a new dataset TABLE2LATEX-450K and proposed to use an attentional encoder-decoder model to convert tables into LaTeX source codes. Li et al. \cite{li2020tablebank} defined a set of HTML tags to describe table structures only and presented a new table benchmark dataset known as TableBank. Zhong et al. \cite{zhong2020image} introduced another large scale table benchmark dataset PubTabNet, which contains 568k table images with corresponding structured HTML representation, and introduced an attention-based encoder-dual-decoder architecture to recognize table structures and cell contents simultaneously. These methods rely on a large amount of training data and still struggle with big and complex tables \cite{zhong2020image,li2020tablebank}.

\textbf{Bottom-up methods.} One group of bottom-up methods \cite{qasim2019rethinking,chi2019complicated,li2021gfte,xue2019res2tim} treat words or cell contents as nodes in a graph and use graph neural networks to predict whether each sampled node pair is in a same cell, row, or column. These methods rely on an assumption that the bounding boxes of words or cell contents are available as additional inputs, which are not easy to obtain from table images directly. To eliminate this assumption, another group of methods \cite{zheng2021global,raja2020table,qiao2021lgpma} proposed to detect the bounding boxes of table cells directly. After cell detection, Zheng et al. \cite{zheng2021global} and Qiao et al. \cite{qiao2021lgpma} designed some rules to cluster cells into rows and columns. In order to improve the accuracy of both cell detection and cell clustering, Raja et al. \cite{raja2020table} introduced a novel loss function that modeled the inherent alignment of cells in the cell detection network, and a graph-based problem formulation to build associations between the detected cells. However, this method still fails to handle tables containing a large number of empty cells and distorted tables.

\section{Methodology}
\label{sec:methodology}
\subsection{Overview}
\label{subsec:overview}
\begin{figure}[t]
    \centering
    \setlength{\abovecaptionskip}{-0.2cm}
    \includegraphics[width=1.0\linewidth]{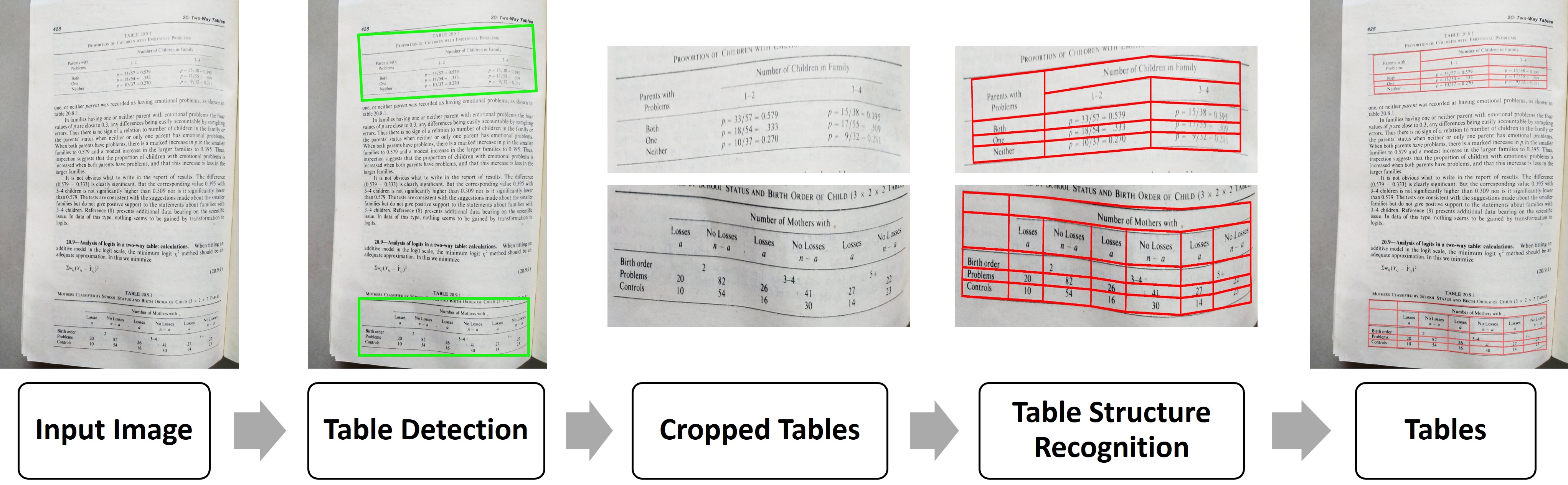}
    \caption{{\color{black}{An outline of our table extraction approach.}}}
    \label{fig:overview}
\end{figure}

Our table extraction approach, RobusTabNet, consists of two deep learning models, i.e., a table detector and a table structure recognizer. {\color{black}{For each input image, we first use our table detector to detect all tables within it and crop them from the original image. Then, each cropped table image is resized to an appropriate resolution and fed into the table structure recognizer to reconstruct its cellular structure. Finally, the recognition results are mapped back onto the original image. An outline of our approach and the expected outputs are shown in Fig.~\ref{fig:overview}.}} Details of our table detector and structure recognizer will be introduced in Section \ref{subsec:CornerNet-FRCN} and Section \ref{subsec:TSR_method}, respectively.

\subsection{CornerNet-FRCN based table detector}
\label{subsec:CornerNet-FRCN}
\begin{figure*}[t]
    \centering
    \setlength{\abovecaptionskip}{-0.2cm}
    \includegraphics[width=1.0\linewidth]{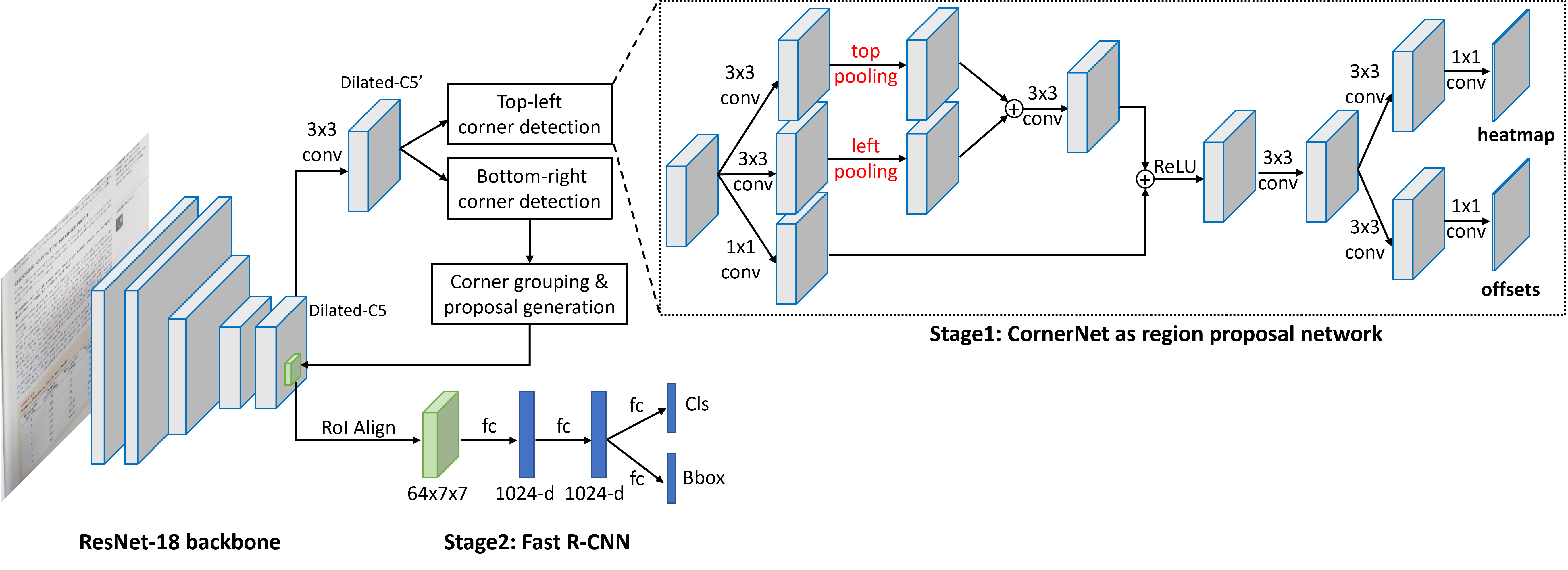}
    \caption{Overall architecture of our CornerNet-FRCN based table detection approach.}
    \label{fig:cornernet_frcn_pipeline}
\end{figure*}
{\color{black}{Existing CNN-based table detection methods typically use RPN to generate table proposals. We find that the percentage of well-localized table proposals (IoU$>$0.9) in the positive samples (IoU$>$0.7) generated by RPN is not high enough, which is an important reason for the unsatisfactory localization accuracy of these table detectors (see analysis in Section~\ref{sebsebsec:abl_corner}). To address this issue, we propose to use CornerNet to detect the top-left and bottom-right corners of all table bounding boxes first and then group each pair of top-left and bottom-right corners to obtain table proposals. As table corners can be precisely inferred from ruling lines and alignment of cell contents in tables, the positive proposals (IoU$>$0.7) generated by our approach will be of higher localization accuracy, which can improve the localization accuracy of our table detector effectively. After that, we use a simple Fast R-CNN module to reject non-table proposals and refine the bounding boxes of remaining positive proposals further.}}

The overall architecture of our approach is illustrated in Fig.~\ref{fig:cornernet_frcn_pipeline}. There are three core modules: 1) A CNN backbone network that is responsible for computing a shared convolutional feature map; 2) A CornerNet based region proposal generation module, which detects the top-left and bottom-right corners of the tables and enumerates all the potential table proposals; 3) A Fast R-CNN (FRCN) module, which is used to prune non-table proposals and refine the bounding boxes of remaining table proposals. For the sake of efficiency, a ResNet-18 network with dilations in “Conv5” is used as the backbone network, and the stride of the output feature map, named \textit{Dilated-C5}, is 16 pixels. We further use a $1\times1$ convolutional layer to reduce the channel dimension of \textit{Dilated-C5} from 512 to 64 for computational efficiency. 

\subsubsection{CornerNet as region proposal network}
\label{subsubsec:CornerNet_as_RPN}
CornerNet \cite{law2018cornernet} detects an object as a pair of keypoints, i.e., the top-left and bottom-right corners of the bounding box.  It uses a convolutional network to predict two sets of heatmaps to represent the locations of the top-left and bottom-right corners of different object categories respectively, as well as an embedding vector for each detected corner such that the distance between the embeddings of two corners from the same object is small. To produce tighter bounding boxes, the network also predicts offsets to slightly adjust the locations of the corners. With the predicted heatmaps, embeddings and offsets, a simple post-processing algorithm is applied to obtain the final bounding boxes. {\color{black}{However, Duan et al. \cite{duan2019centernet} find that the performance of the abovementioned corner grouping method is restricted by its relatively weak ability of referring to the global information of an object. Therefore, in this work, we abandon the embedding vectors and adopt CornerNet as a new region proposal network for Faster R-CNN by detecting and exhaustively grouping corner points.}}

As illustrated in Fig.~\ref{fig:cornernet_frcn_pipeline}, we append a $3\times3$ convolutional layer with the stride of 1 on \textit{Dilated-C5} to generate a new feature map \textit{Dilated-C5'}, on which two sibling branches are attached for detecting top-left and bottom-right corners, respectively. Taking the top-left corner detection branch as an example, we first use a top-left corner pooling module, composed of a top pooling and a left pooling operator \cite{law2018cornernet}, to aggregate context information. The context enhanced feature map is fused with the original feature map in a residual connection manner. Then, a detection module is attached to this feature map and performs dense per-pixel prediction of top-left corners. Specifically, let $p_i$ and $q_j$ be a pixel on the feature map and raw image with the coordinates of $(p_i^x,p_i^y)$ and $(q_j^x,q_j^y)$, respectively. We define that $p_i$ is corresponding to $q_j$ if
\begin{equation}
    p_i^x=\left\lfloor \frac{q_j^x}{s} \right\rfloor and~ p_i^y=\left\lfloor \frac{q_j^y}{s} \right\rfloor,
    \label{equ:point_corresponging}
\end{equation}
where $s$ denotes the stride of the feature map. If $p_i$ is corresponding to $q_j$ and $q_j$ is a top-left corner point, the detection module will give $p_i$ a “top-left corner” label and predict the corresponding offset $\Delta_i$ defined by
\begin{equation}
    \Delta_i=\left(\frac{q_j^x}{s} - \left\lfloor \frac{q_j^x}{s} \right\rfloor, \frac{q_j^y}{s} - \left\lfloor \frac{q_j^y}{s} \right\rfloor\right),
    \label{equ:corner_offset}
\end{equation}
to adjust the location of the corner to compensate for the quantization error caused by network down-sampling. As depicted in Fig.~\ref{fig:cornernet_frcn_pipeline}, the detection module contains two parallel branches, a $3\times3$ convolutional layer followed by a $1\times1$ convolutional layer in each branch, for corner/non-corner classification and corner offset regression, respectively. Furthermore, if a pair of false corner detections are close to the corresponding ground truth corner locations, they can still produce a box that highly {\color{black}{overlaps}} with the ground-truth box. Therefore, during training, we reduce the penalty of the negative locations within a radius $r$ of the positive location, and the radius $r$ is determined by the size of the ground-truth box. We refer readers to \cite{law2018cornernet} for more details about the selection of $r$. Once $r$ is determined, the amount of penalty reduction is given by an unnormalized 2D Gaussian, $e^{-\frac{x^2+y^2}{2\sigma^2}}$, whose center is at the positive location and $\sigma$ is set as $r/3$. 

To generate table proposals, we first apply non-maximal suppression (NMS) by using a $3\times3$ max pooling layer on the corner heatmaps. {\color{black}{Then top-$K$ top-left and bottom-right corners are extracted from the heatmaps, which are further filtered by a score threshold, $C_{th}$.}}
The locations of the remaining corners are adjusted by the corresponding predicted offsets. 
Then, we take all the valid combinations, i.e., the $x$ and $y$ coordinates of the top-left corner are smaller than that of the bottom-right corner, as table proposals, so that a high recall rate is retained. After that, we use the standard NMS algorithm with an IoU threshold of 0.7 to remove redundant proposal boxes.

\subsubsection{Fast R-CNN}
\label{subsubsec:FRCN}
Given the extracted region proposals, we adopt a Fast R-CNN module to reject negative (non-table) proposals and refine the bounding boxes of positive (table) proposals. As shown in Fig.~\ref{fig:cornernet_frcn_pipeline}, for each proposal, we first adopt an RoI Align algorithm \cite{he2017mask} to extract a $7\times7\times64$ feature descriptor from the proposal box on the \textit{Dilated-C5} feature map. Then, it is fed into two 1,024-d fully connected ($fc$) layers (each followed by a ReLU activation function) before the final table/non-table classification and bounding box regression layers. During training, a proposal is assigned a positive label if it has an IoU over 0.7 with any ground-truth bounding box, or a negative label if it has IoU lower than 0.5 for all ground-truth bounding boxes. An online hard example mining (OHEM) method is adopted to select an equal number of hard positive and hard negative samples to train the Fast R-CNN module.

\begin{figure}[t]
    \centering
    \setlength{\abovecaptionskip}{-0.2cm}
    \includegraphics[width=1.0\linewidth]{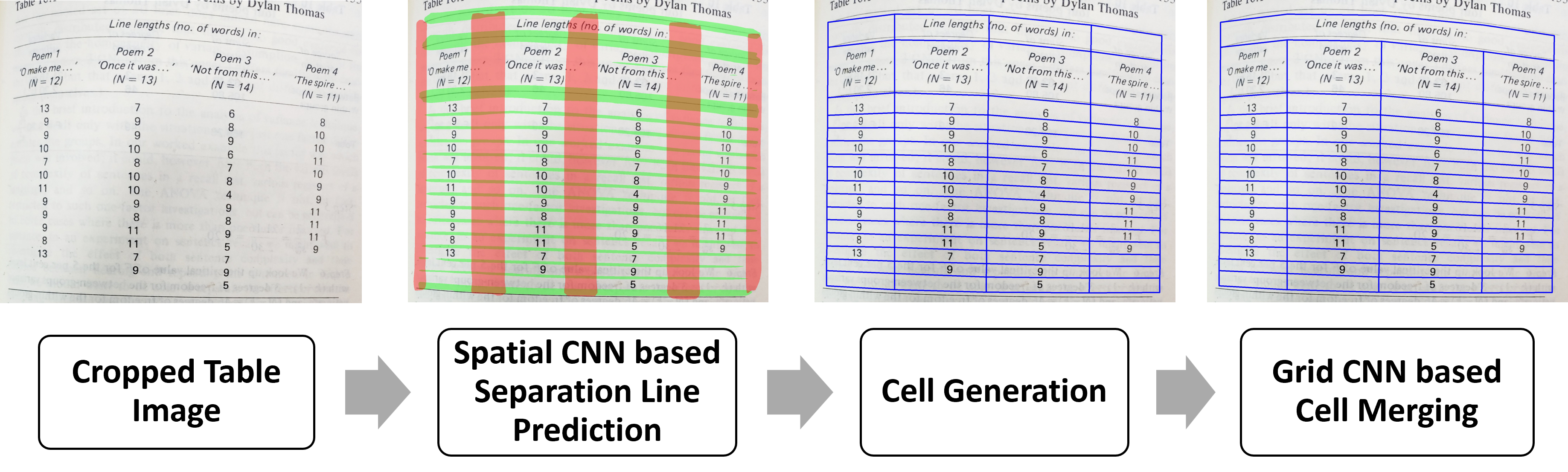}
    \caption{Flowchart of our table structure recognition approach.}
    \label{fig:Flowchart_of_TSR}
\end{figure}

\subsection{Split-and-merge based table structure recognizer}
\label{subsec:TSR_method}
After table detection, each detected table is cropped from the raw image and resized to an appropriate size {\color{black}{to ensure that there is enough inter-line spacing for separation line prediction.}} Then, each resized table image is fed into a table structure recognizer to reconstruct its cellular structure. 
The flowchart of our table structure recognizer is shown in Fig.~\ref{fig:Flowchart_of_TSR}. Given a cropped table image, a spatial CNN based separation line prediction module is used to predict a row separator mask and a column separator mask first. Then, a connected component analysis (CCA) based line generation algorithm is used to extract all row and column separation lines from the predicted separator masks, which are intersected to generate a grid of cells. After that, a Grid CNN based cell merging module is adopted to merge wrongly split cells into spanning cells. The separation line prediction module and cell merging module share a same CNN backbone network and are trained jointly. For the sake of efficiency, we adopt the Feature Pyramid Network (FPN) \cite{lin2017feature}, which is built on top of ResNet-18, as the backbone network. Details of the separation line prediction and cell merging modules are described in Section \ref{subsubsec:SCNN} and \ref{subsubsec:GridCNN}, respectively.

\begin{figure*}[t]
    \centering
    \setlength{\abovecaptionskip}{-0.2cm}
    \includegraphics[width=1.0\linewidth]{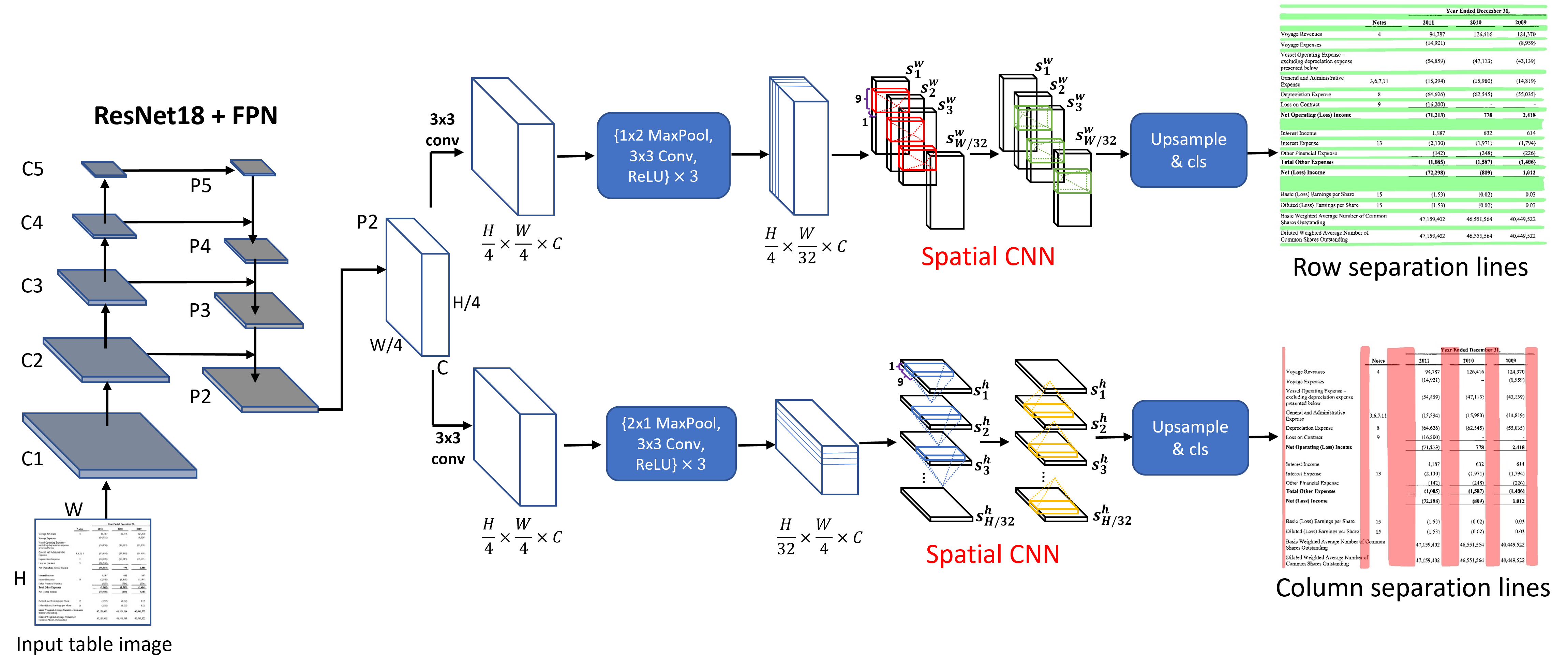}
    \caption{Overall architecture of our spatial CNN based separation line prediction module.}
    \label{fig:SCNN}
\end{figure*}
\subsubsection{Spatial CNN based separation line prediction} 
\label{subsubsec:SCNN}
{\color{black}{Some important visual clues like ruling lines and alignment of cell contents provide useful hints to indicate whether a separation line exists at a position within a table. However, the ResNet-FPN backbone cannot embed such useful visual clues into the features of pixels in large blank regions of borderless tables effectively, because each feature vector on the output convolutional feature map only contains local context information extracted from its effective receptive field. Based on such convolutional feature map, it is hard for the following separation line segmentation module to predict separation lines from large blank spaces in borderless tables robustly, because the feature vector of each pixel in these blank regions does not contain enough information to determine whether a separator line passes through this pixel or not. To address this issue, we propose to use spatial CNN modules \cite{pan2018spatial} to enhance the feature representation of each pixel on the convolutional feature map by propagating contextual information across the whole feature map in left-right or top-bottom directions.}}

The overall architecture of our spatial CNN based separation line prediction module is depicted in Fig.~\ref{fig:SCNN}. Given an input image $X \in R^{H \times W \times 3}$, we adopt the FPN backbone network to generate a shared convolutional feature map $P_2 \in R^{\frac{H}{4} \times \frac{W}{4} \times C}$, where $C$ represents the number of channels and is set to 64 in our experiments. Then, two parallel semantic segmentation branches are attached to $P_2$ to predict a row separator mask $\hat{S}^{row}$ and a column separator mask $\hat{S}^{col}$, respectively. Taking the row separation line prediction branch as an example, we add a $3\times3$ convolutional layer and three repeated down-sampling blocks, each composed of a sequence of a $1\times2$ max-pooling layer, a $3\times3$ convolutional layer and a ReLU activation function, after $P_2$ sequentially to generate a down-sampled feature map $P_2^{'} \in R^{\frac{H}{4} \times \frac{W}{32} \times C}$, which is taken as the input of two cascaded spatial CNN modules. The first spatial CNN module divides the feature map into $\frac{W}{32}$ slices along the width direction, which are denoted as $S^w = \{s_i^w \in R^{\frac{H}{4} \times 1 \times C} | i \in N, i = 1,2,...,\frac{W}{32}\}$ then propagates the information from the leftmost slice $s_1^w$ to the rightmost slice $s_{W⁄32}^w$ with convolution operators. Specifically, the leftmost slice $s_1^w$ is convolved by a convolution kernel with the kernel size of $9\times1$ (9 and 1 represent kernel height and width respectively) and its output feature map is merged with its right slice $s_2^w$ by element-wise addition. This procedure is done iteratively so that the information can be propagated from the leftmost slice to the rightmost slice effectively. The second spatial CNN module uses the same method to propagate the information from the rightmost slice $s_{W⁄32}^w$ to the leftmost slice $s_1^w$. In this way, each pixel in the output feature map can leverage the structural information from both sides to enhance its feature representation ability. Finally, this context-enhanced feature map is up-sampled by a factor of 4 with a bilinear interpolation operation to generate an output feature map $P_{out} \in R^{H \times \frac{W}{8} \times C}$, on which a $1\times1$ convolutional layer followed by a sigmoid activation function is attached to predict a row separator mask $\hat{S}^{row} \in R^{H \times \frac{W}{8} \times 1}$. The architecture of the column separation line prediction branch is similar to the row separation line prediction branch, except that the down-sampling is performed along the height direction and the two spatial CNN modules propagate information from the topmost slice to the bottommost slice and from the bottommost slice to the topmost slice, respectively.

\begin{figure}[t]
    \centering
    \setlength{\abovecaptionskip}{-0.3cm}
    \includegraphics[width=1.0\linewidth]{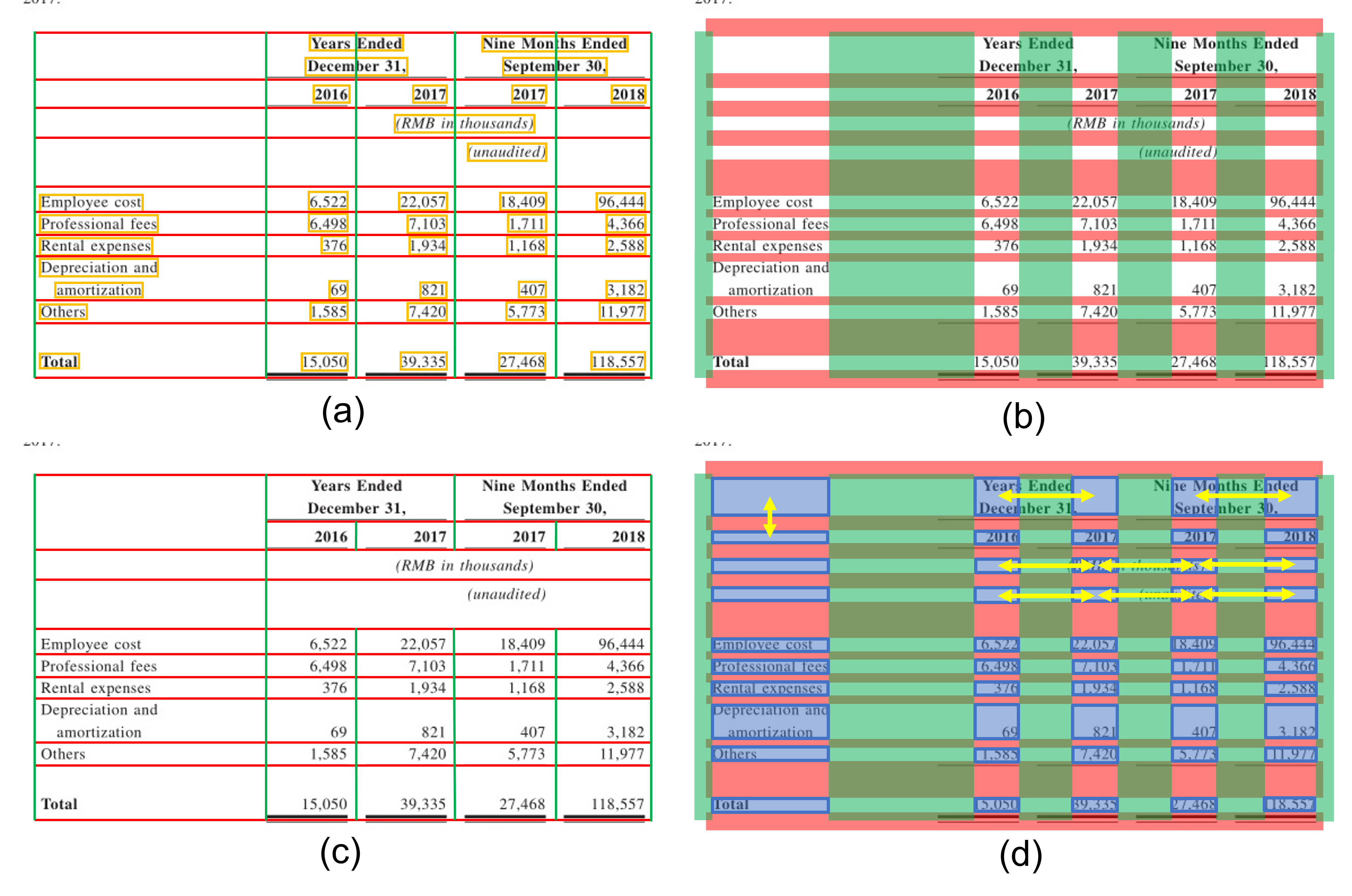}
    \caption{Illustration of the ground-truth generation for table structure recognition. (a) Annotated text boxes and separation lines; (b) Expanded separation lines; (c) Ground-truth cell boxes, including spanning cells; (d) If a pair of neighboring shrunk cells (blue boxes) detected by the split model are assigned to a same ground-truth cell box, we will give this pair a positive label, otherwise a negative label, to train the cell merging module. (For interpretation of the references to colour in this figure legend, the reader is referred to the web version of this article.)}
    \label{fig:gt_generation}
\end{figure}

To generate the ground-truth (GT) row and column separator masks, the row and column separation lines of each table as well as the bounding boxes of text-lines in each cell are annotated (Fig.~\ref{fig:gt_generation}(a)). Following SPLERGE \cite{tensmeyer2019deep}, we calculate the GT separator masks by maximizing the size of the separation regions without intersecting any non-spanning cell contents, as shown in Fig.~\ref{fig:gt_generation}(b). Specifically, for each annotated row separation line, we move it upwards and downwards respectively until it touches a text box that belongs to a non-spanning cell to obtain its corresponding row separation region. The similar procedure can also be applied to generate column separation regions. After this step, if the thickness of a separation region is less than 8 pixels, we will further expand it to make sure that its thickness is at least 8 pixels.

\subsubsection{Cell generation} 
\label{subsubsec:cell_generation}
\begin{figure}[t]
    \centering
    \setlength{\abovecaptionskip}{-0.2cm}
    \includegraphics[width=1.0\linewidth]{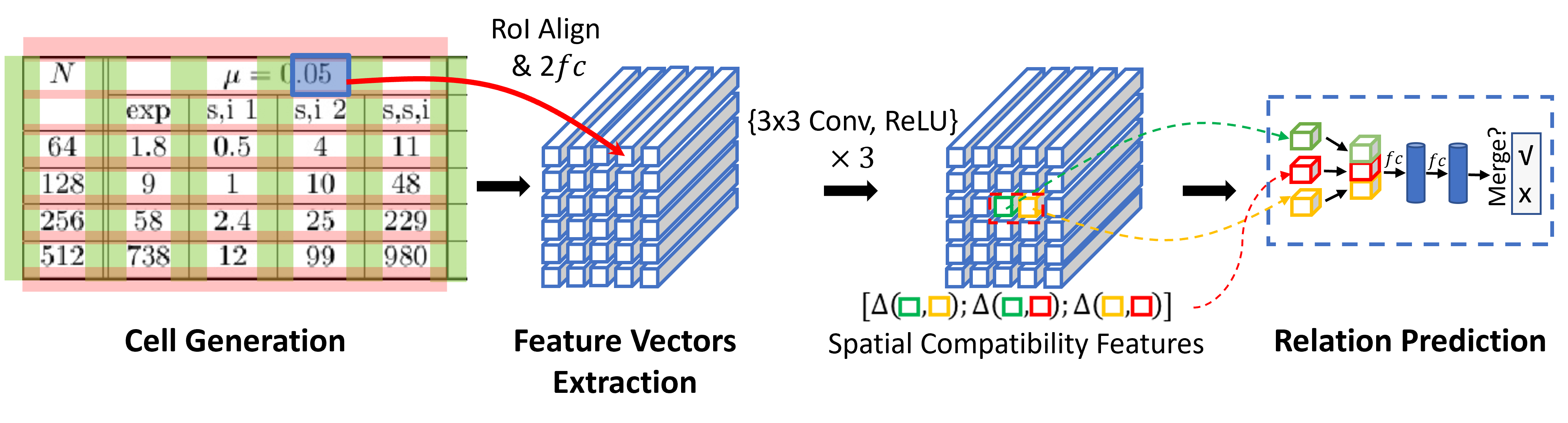}
    \caption{A schematic view of the Grid CNN based cell merging module.}
    \label{fig:GridCNN}
\end{figure}

{\color{black}{After separation line prediction, we first binarize the predicted row and column separator heatmaps with a classification score threshold, $S_{th}$.}} Then, we extract the connected components (CCs) from the segmentation masks, which represent detected separators. With these CCs, we can extract all row and column separation lines as well as their corresponding line thicknesses. Taking row separation line generation as an example, we first apply the \textit{findContours} method in OpenCV \cite{bradski2000opencv} to the binarized row separator mask to obtain the contours of row CCs. Then for each row CC, we use a polynomial curve fitting algorithm to fit a function $y=f(x)$ from its contour points, which approximates the center line of this row CC. To compute the corresponding line thickness, a vertical scan-line is used to traverse the related row CC mask from left to right with a stride of 8 pixels. On each scanned pixel column, a line segment can be obtained by intersecting the scan-line with the upper and lower boundaries of the row CC. By averaging the lengths of all the line segments, we can estimate the line thickness of the separation line, denoted by $lw$. Then, we translate the fitted separation line $y=f(x)$ upwards and downwards respectively to generate two border lines, $y=f(x)+\frac{lw}{2}$ and $y=f(x)-\frac{lw}{2}$. Similarly, the column separation lines can also be generated by rotating the column separator heatmap with 90 degrees before running this algorithm. Finally, we intersect all the translated row lines with column lines to calculate all intersection points, from which we can extract all the shrunk cell boxes, e.g., the blue box in Fig.~\ref{fig:GridCNN}, and arrange them in a grid manner.

\subsubsection{Grid CNN based cell merging} 
\label{subsubsec:GridCNN}
{\color{black}{ Based on the compact grid representation, we introduce a Grid CNN module to aggregate context information effectively with several stacked convolution layers to improve cell merging accuracy.}} As shown in Fig.~\ref{fig:GridCNN}, we first use an RoI Align algorithm to extract a $7 \times 7 \times 64$ feature descriptor from the bounding box of each cell, which is then fed into a 2-hidden-layer fully connected ($fc$) neural network with 512 nodes at each layer to generate a 512-d feature vector. {\color{black}{Assume that the detected cells are arranged in $M$ rows and $N$ columns, then the corresponding feature vectors will construct a new feature map $\boldsymbol{F_{grid}} \in R^{M \times N \times 512}$.}} Each pixel in $\boldsymbol{F_{grid}}$ corresponds to a cell generated by the split model. $\boldsymbol{F_{grid}}$ is then convolved by three $3\times3$ convolutional layers to obtain an enhanced feature map $\boldsymbol{F_{grid}^{'}} \in R^{M \times N \times 512}$. Finally, we use a relation network \cite{zhang2017relationship} to predict whether each pair of adjacent pixels on $\boldsymbol{F_{grid}^{'}}$, which corresponds to each pair of adjacent cells in the input table image, should be merged or not. Here, we only consider 4-adjacency neighborhood relations to construct relational pairs.

The architecture of the relation network is shown in the blue dashed box in Fig.~\ref{fig:GridCNN}. Given a pair of adjacent pixels in $\boldsymbol{F_{grid}^{'}}$, $p_i$ and $p_j$, whose corresponding feature vectors are $\boldsymbol{F_{grid}^{'p_i}}$ and $\boldsymbol{F_{grid}^{'p_j}}$ and corresponding cell bounding boxes are $b_i$ and $b_j$ respectively, we extract a feature representation $\boldsymbol{x_{ij}}$ to encode the appearance compatibility and spatial compatibility of their corresponding cells. Specifically, $\boldsymbol{x_{ij}}$ is constructed by concatenating the appearance features $\boldsymbol{F_{grid}^{'p_i}}$ and $\boldsymbol{F_{grid}^{'p_j}}$ and the spatial compatibility feature $\boldsymbol{l_{ij}}$ of $b_i$ and $b_j$, i.e., $\boldsymbol{x_{ij}}=[\boldsymbol{F_{grid}^{'p_i }};\boldsymbol{l_{ij}};\boldsymbol{F_{grid}^{'p_j}}]$. The spatial compatibility feature $\boldsymbol{l_{ij}}$ is used to measure the relative scale and location relationships between $b_i$ and $b_j$. Following Zhang et al. \cite{zhang2017relationship}, let $b_{ij}$ denote the union bounding box of $b_i$ and $b_j$, then $\boldsymbol{l_{ij}}$ is defined as an 18-d vector concatenating three 6-d vectors, which indicate the box delta of $b_i$ and $b_j$, $b_i$ and $b_{ij}$, $b_j$ and $b_{ij}$, respectively. Given two bounding boxes $b_i=\{x^i,y^i,w^i,h^i\}$ and $b_j=\{x^j,y^j,w^j,h^j\}$, their box delta is defined as $\Delta(b_i,b_j )=(t_x^{ij},t_y^{ij},t_w^{ij},t_h^{ij},t_x^{ji},t_y^{ji})$ where each dimension is given by
\begin{align}
    \nonumber
    &t_{x}^{ij}=\left(x^{i}-x^{j}\right) / w^{i}, \quad t_{y}^{ij}=\left(y^{i}-y^{j}\right) / h^{i}, \\ 
    &t_{w}^{ij}=\log \left(w^{i} / w^{j}\right), \quad 
    t_{h}^{ij}=\log \left(h^{i} / h^{j}\right), \\
    \nonumber
    &t_{x}^{ji}=\left(x^{j}-x^{i}\right) / w^{j}, \quad t_{y}^{ji}=\left(y^{j}-y^{i}\right) / h^{j}.
\end{align}
A binary classifier is applied on the feature representation to predict whether each pair of cells should be merged or not. It is implemented with a 2-hidden-layer MLP with 512 nodes at each hidden layer and a sigmoid activation node at its output layer. Note that in the inference stage, each pair of cells is predicted twice for the inputs $\boldsymbol{x_{ij}}$ and $\boldsymbol{x_{ji}}$, and the maximum value is taken as the final merging score for this pair of cells. 

In the training stage, we use detected cells from the split model to generate positive and negative relational pairs for training the cell merging module. As illustrated in Fig.~\ref{fig:gt_generation}(c-d), given all the ground-truth cell boxes, a detected cell box $\boldsymbol{b_{det}}$ is assigned to a ground-truth cell box $\boldsymbol{b_{gt}}$ if the following condition is satisfied, i.e.,
\begin{equation}
    \frac{Area(\boldsymbol{b_{det}} \cap \boldsymbol{b_{gt}})}{Area(\boldsymbol{b_{det}})}>0.5,
\end{equation}
where $Area(\boldsymbol{b_{det}})$ and $Area(\boldsymbol{b_{det}} \cap \boldsymbol{b_{gt}})$ denote the area of $\boldsymbol{b_{det}}$ and the area of the overlap between $\boldsymbol{b_{det}}$ and $\boldsymbol{b_{gt}}$, respectively. Then, each detected cell is paired with each of its 4-connected cells to construct candidate relational pairs. If two cells in a relational pair are assigned to a same ground-truth cell box, we give this relational pair a positive label, otherwise a negative label. During training, we ignore all the negative relational pairs that contain cells not assigned to any ground-truth cell, and then adopt an OHEM method to select hard samples to train the cell merging module.

\section{Loss Function}
\label{subsec:loss_func}
\subsection{Table detection}
\textbf{Loss for CornerNet based region proposal network.} 
There are two sibling output layers for each corner detection branch, i.e., a corner/non-corner classification layer and an offset regression layer. The multi-task loss function can be defined as follows:
\begin{equation}
    \mathcal{L}_{corner} = \frac{1}{N_t}\sum_{i}\mathcal{L}_{det}(c_i, c_i^*) + \frac{1}{N_{c}}\sum_{j}\mathcal{L}_{off}(t_j, t_j^*),
    \label{equ:cornernet_loss}
\end{equation}
where $N_t$ and $N_c$ denote the number of tables and the number of corners in a mini-batch respectively, $c_i$ and $c_i^*$ are the predicted and “ground-truth” labels for the $i$-th pixel on the heatmap, $c_i^*$ has been augmented with the unnormalized Gaussians to reduce the penalty around the ground-truth locations, $\mathcal{L}_{det}(c_i,c_i^*)$ is a variant of focal loss as in \cite{law2018cornernet} for classification tasks, $t_j$ and $t_j^*$ are predicted and ground-truth 2-d coordinate offsets defined by Eq.~\ref{equ:point_corresponging} and Eq.~\ref{equ:corner_offset} for the $j$-th corner, $\mathcal{L}_{off} (t_j,t_j^*)$ is a Smooth-$L_1$ loss\cite{ren2015faster} for regression tasks.

\textbf{Loss for Fast R-CNN.} There are two sibling output layers for the Fast R-CNN module, i.e., a table/non-table classification layer and a quadrilateral bounding box regression layer. The multi-task loss function is defined as follows: 
\begin{equation}
    \mathcal{L}_{frcn} = \frac{1}{N}\sum_{i}\mathcal{L}_{cls}(k_i,k_i^*) + \frac{1}{N_{fg}}\sum_{j}\mathcal{L}_{reg}(b_j, b_j^*),
\end{equation}
where $N$ is the number of sampling region proposals (including $N_{fg}$ positive ones), $k_i$ and $k_i^*$ are predicted and ground-truth labels for the $i$-th sampling region proposal respectively, $\mathcal{L}_{cls}(k_i,k_i^*)$ is a cross-entropy loss for classification tasks, $b_j$ and $b_j^*$ are predicted and ground-truth 8-d normalized coordinate offsets as stated in \cite{zhong2019anchor} for the $j$-th positive region proposal, $\mathcal{L}_{reg}(b_j,b_j^*)$ is an $L_1$ loss for regression tasks.

\textbf{Total loss for table detector.} With the definitions of $\mathcal{L}_{corner}$ and $\mathcal{L}_{frcn}$, the training loss for the table detector can be defined as follows:
\begin{equation}
   \mathcal{L}_{detector} = \lambda_{corner} \cdot \mathcal{L}_{corner}+\mathcal{L}_{frcn},
\end{equation}
where $\lambda_{corner}$ is a loss-balancing parameter, and we set $\lambda_{corner}=0.2$.

\subsection{Table structure recognition}
\textbf{Loss for spatial CNN based separation line prediction.} There are two branches in the separation line prediction module for row and column separator prediction, respectively. The total loss of this module is the sum of the losses of two branches. Let $N_{row}$ and $N_{col}$ denote the number of sampling pixels for row and column separator prediction branch respectively, $\{R_i,C_j\}$ and $\{R_i^*,C_j^*\}$ be the predicted and ground-truth labels for the $i$-th sampling pixel on the row separator heatmap and the $j$-th sampling pixel on the column separator heatmap respectively, and $\mathcal{L}(R_i,R_i^*)$ and $\mathcal{L}(C_j,C_j^*)$ be the binary cross-entropy loss for classification tasks. Based on these definitions, the loss function for the separation line prediction module can be defined as follows:
\begin{equation}
    \mathcal{L}_{split} = \frac{1}{N_{row}}\sum_i\mathcal{L}(R_i,R_i^*)+\frac{1}{N_{col}}\sum_j\mathcal{L}(C_j,C_j^*).
\end{equation}

\textbf{Loss for Grid CNN based cell merging.} Let $N_p$ be the number of selected relational pairs for cell merging, $r_i$ and $r_i^*$ be the predicted and ground-truth labels for the $i$-th relational pair, and $\mathcal{L}(r_i,r_i^*)$ be a binary cross-entropy loss for classification tasks. The loss function for the cell merging module is defined as follows:
\begin{equation}
    \mathcal{L}_{merge} = \frac{1}{N_p}\sum_i\mathcal{L}(r_i,r_i^*).
\end{equation}

\textbf{Total loss for table structure recognizer.} With the definitions of $\mathcal{L}_{split}$ and $\mathcal{L}_{merge}$, the training loss for the table structure recognizer can be defined as follows:
\begin{equation}
   \mathcal{L}_{recognizer} = \mathcal{L}_{split}+\mathcal{L}_{merge}.
\end{equation}

\section{Experiments}
\label{sec:exps}
\subsection{Datasets and evaluation protocols}
\label{subsec:data}
We conduct comprehensive experiments on three table detection benchmark datasets, including cTDaR TrackA \cite{gao2019icdar}, PubLayNet \cite{zhong2019publaynet} and IIIT-AR-13K \cite{mondal2020iiit}, and three table structure recognition datasets, including SciTSR \cite{chi2019complicated}, PubTabNet \cite{zhong2020image} and {\color{black}{cTDaR TrackB2-Modern}} \cite{gao2019icdar}, to evaluate the performance of our table detection and structure recognition approaches, respectively. We follow the evaluation protocols defined by the authors to make our results comparable to the ones reported by other methods. Moreover, to demonstrate the advantage of our TSR approach in dealing with geometrically distorted tables, we have also collected a much more challenging in-house dataset which contains many distorted or even curved tables.

\textbf{cTDaR TrackA} \cite{gao2019icdar} contains both historical and modern document images. The historical subset contains hand-drawn tables and handwritten texts, including 600 images for training and 199 images for testing. The modern subset contains printed PDF documents, including 600 images for training and 240 images for testing. It adopts the weighted average (WAvg.) F1-score as evaluation metric, which is calculated with the IoU thresholds of 0.6, 0.7, 0.8 and 0.9.

\textbf{PubLayNet} \cite{zhong2019publaynet} is a high-quality dataset for document layout analysis, which contains 335,703 images for training, 11,245 images for validation and 11,405 images for testing. We use it for table detection performance evaluation, and only use the images containing at least a table for model training (86,460 images). Since the annotations of the testing set are not released, we only report results on the validation set. The COCO evaluation protocol is used as the evaluation metric of this dataset.

\textbf{IIIT-AR-13K} \cite{mondal2020iiit} is introduced for graphical object detection in annual reports, which contains 9,333 images for training, 1,955 images for validation and 2,120 images for testing. This dataset is used for table detection performance evaluation only. The PASCAL VOC evaluation protocol is used as the evaluation metric of this dataset. 

\textbf{SciTSR} \cite{chi2019complicated} contains 12,000 training images and 3,000 testing images cropped from scientific papers. To evaluate the performance of different methods on complicated tables, authors also extract all the 716 complicated tables from the test set as a test subset, called SciTSR-COMP. The adjacency relation-based evaluation metric, which is used in ICDAR-2013 table competition \cite{gobel2013icdar}, is employed as the evaluation metric of this dataset.

\textbf{PubTabNet} \cite{zhong2020image} contains 500,777 training images, 9,115 validating images and 9,138 testing images. This dataset contains a large number of three-lines tables with empty or spanning cells. Since the annotations of testing set are not released, we only report results on the validation set. The authors proposed a new Tree-Edit-Distance-based Similarity (TEDS) metric for table recognition task, which can identify both table structure recognition and OCR errors. Some recent works \cite{zheng2021global,raja2020table,qiao2021lgpma} have proposed a modified TEDS metric, denoted as TEDS-Struct, to evaluate table structure recognition accuracy only by ignoring OCR errors. We also use the TEDS-Struct metric to evaluate our table structure recognition approach on this dataset.

\textbf{cTDaR TrackB2-Modern} \cite{gao2019icdar} contains no images for training, but 100 images with annotations are provided as testing data. To evaluate our approach on this dataset, we manually labeled the structures of tables in the cTDaR TrackA modern subset, which contains 600 training images. The annotations will be released publicly to facilitate future research in this area. It has been checked that there is no overlap between the 600 training images and the 100 testing images. The adjacency relation-based metric\footnote{https://github.com/cndplab-founder/ctdar\_measurement\_tool} is used as the evaluation metric of this dataset. During evaluation, the convex hull of the content is used to represent a cell. Note that both table region detection and table structure recognition have to be done on this dataset.

\textbf{Private Dataset.} Our in-house dataset is composed of 9,000 training images and 700 testing images. Most images in this dataset are captured by cameras so that many tables in this dataset are skewed or even curved. Sample images in this dataset are shown in Fig.~\ref{fig:TSR_curved_demo} and Fig.~\ref{fig:scnn_demo}. We use the same adjacency relation-based metric as cTDaR TrackB to evaluate our table structure recognition approach on this dataset.

\subsection{Implementation Details}
\label{subsec:details}
For both table detector and structure recognizer, the weights of ResNet-18 related layers are initialized with a pre-trained ResNet-18 model for the ImageNet classification task. The weights of newly added layers are initialized with a Gaussian distribution of mean 0 and standard deviation 0.01. The models are optimized by a standard SGD algorithm with a momentum of 0.9 and weight decay of 0.0005. Unless otherwise specified, all the models are trained for 15K iterations with a base learning rate of 0.032, which is divided by 10 at 10K and 13K iterations, respectively. The table detection model for PubLayNet is trained with a $3\times$ training schedule because of the larger amount of data. The TSR models for SciTSR and PubTabNet are trained for 12 epochs. Besides, we apply synchronized batch normalization across multiple GPUs to stabilize the training.   

\begin{table*}[t]
    \setlength{\tabcolsep}{2.9pt}
    \footnotesize
    \centering
    \caption{Table detection performance comparison on ICDAR2019 cTDaR TrackA. * indicates that the results are from \cite{gao2019icdar}}
    \label{tab:cTDaR_TrackA}
    \begin{tabular}{ c  c  c  c  c  c  c  c  c  c  c  c  c  c  c  c  c}
        \toprule
        \multirow{2}{*}{Methods} & \multicolumn{3}{c}{IoU@0.6(\%)} && \multicolumn{3}{c}{IoU@0.7(\%)} &&
        \multicolumn{3}{c}{IoU@0.8(\%)} &&
        \multicolumn{3}{c}{IoU@0.9(\%)} & WAvg. \\\cline{2-4}\cline{6-8}\cline{10-12}\cline{14-16}
         & P & R & F1 && P & R & F1 && P & R & F1 && P & R & F1 & F1(\%)\\
        \midrule
        Applica-robots$^*$ & 90.3 & 90.1 & 90.2 && 88.4 & 88.1 & 88.2 && 82.6 & 82.4 & 82.5 && 54.6 & 54.4 & 54.5 & 77.0  \\
        ABC Fintech$^*$ & 87.4 & 78.5 & 82.7 && 86.3 & 77.5 & 81.7 && 84.1 & 75.5 & 79.6 && 76.8 & 69.0 & 72.7 & 78.6 \\
        Lenovo Ocean$^*$ & 91.8 & 90.2 & 90.1 && 90.8 & 89.2 & 90.0 && 88.5 & 87.0 & 87.7 && 82.9 & 81.5 & 82.2 & 87.7  \\
        NLPR-PAL$^*$ & 97.1 & 97.5 & \textbf{97.3} && 96.0 & 96.4 & \textbf{96.2} && 93.6 & 94.0 & 93.8 && 86.5 & 86.9 & 86.7 & 92.9  \\
        TableRadar$^*$ & 97.6 & 96.4 & 97.0 && 96.6 & 95.4 & 96.0 && 95.8 & 93.2 & 95.1 && 90.8 & 89.7 & 90.2 & 94.2 \\
        CDeC-Net\cite{agarwal2021cdec} & 98.0 & 93.9 & 95.9 && 97.7 & 93.6 & 95.6 && 97.1 & 93.0 & 95.0 && 93.4 & 89.5 & 91.5 & 94.3 \\
        \midrule
        \textbf{RPN+FRCN} & 97.8 & 94.8 & 96.3 && 97.3 & 94.3 & 95.7 && 96.3 & 93.3 & 94.7 && 92.4 & 89.5 & 90.9 & 94.1 \\
        \textbf{Ours (CornerNet+FRCN)} & 98.4 & 94.0 & 96.1 && 98.2 & 93.9 & 96.0 && 97.7 & 93.3 & \textbf{95.4} && 95.0 & 90.8 & \textbf{92.9} & \textbf{94.9} \\
        \bottomrule
    \end{tabular}
\end{table*}
\begin{table*}[t]
    \footnotesize
    \centering
    \caption{Table detection performance comparison on the validation set of PubLayNet. * indicates that the results are from \cite{zhong2019publaynet}.} 
    \label{tab:PubLayNet}
    \color{black}{
    \begin{tabular}{ c  c  c  c  c}
        \toprule
        Methods & Backbone & $AP^{0.5:0.95}$ & $AP^{0.75}$ & $AP^{0.95}$\\
        \midrule
        Faster R-CNN$^*$ & ResNeXt-101 & 95.4 & 97.8 & 77.8 \\
        Mask R-CNN$^*$ & ResNeXt-101 & 96.0 & 97.8 & 81.4 \\
        CDeC-Net\cite{agarwal2021cdec} & Dual ResNeXt-101 & 96.7 & - & - \\
        \midrule
        \textbf{RPN+FRCN} & ResNet-18 & 96.0 & 97.5 & 87.0 \\
        \textbf{Ours (CornerNet+FRCN)} & ResNet-18 & \textbf{97.0} & \textbf{97.8} & \textbf{92.0} \\
        \bottomrule
    \end{tabular}
    }
\end{table*}
\begin{table*}[h!]
    \footnotesize
    \centering
    \caption{Table detection performance comparison on IIIT-AR-13K. * indicates that the results are from \cite{mondal2020iiit}.}
    \label{tab:IIIT-AR-13K}
    \begin{tabular}{ c  c  c  c  c  c  c  c  c  c  c }
        \toprule
        \multirow{2}{*}{Methods} &
        \multirow{2}{*}{Backbone} & \multicolumn{4}{c}{Validation Set(\%)} && \multicolumn{4}{c}{Testing Set(\%)} \\\cline{3-6}\cline{8-11}
         && P & R & F1 & AP && P & R & F1 & AP\\
        \midrule
        Faster R-CNN$^*$ & ResNet-101 & 95.7 & 92.6 & 94.2 & 95.5 && 95.1 & 92.3 & 93.7 & 93.9 \\
        Mask R-CNN$^*$ & ResNet-101 & 98.2 & 96.6 & 97.4 & 97.6 && 97.1 & 97.1 & 97.1 & 96.5 \\
        \midrule
        \textbf{Ours (CornerNet+FRCN)} & \textbf{ResNet-18} & \textbf{98.6} & \textbf{98.3} & \textbf{98.5} & \textbf{98.2} && \textbf{99.0} & \textbf{97.8} & \textbf{98.4} & \textbf{97.7} \\
        \bottomrule
    \end{tabular}
\end{table*}

We implement our approach based on PyTorch\footnote{\url{https://pytorch.org/}} v1.6.0 and conduct experiments on a workstation with 8 Nvidia V100 GPUs. In each training iteration, we sample 4 images for each GPU. In the training phase of our table detector, since the number of positive proposals is small, we add some synthesized samples by introducing random jittering to ground-truth boxes. Then, for each image, we select a mini-batch of 32 hard positive and 32 hard negative proposals for the FRCN detector. We adopt a multi-scale training strategy during training. While keeping the aspect ratio, the shorter side of each selected training image is randomly rescaled to a number in $\{320, 416, 512, 608, 704, 800\}$. Moreover, when training our table detector on cTDaR TrackA, we also rotate training images by a random angle in $\{0^\circ, 90^\circ, 180^\circ, 270^\circ\}$ with $\pm5^\circ$ angle jitter for data augmentation. In the training phase of our table structure recognizer, we use the cropped table images for training and the shorter side of each selected training image is randomly rescaled to a number in $\{416, 512, 608, 704, 800\}$ while keeping the aspect ratio. For each image, we sample a mini-batch of 1,024 row/column separator pixels and 1,024 background pixels for each separation line prediction branch. Furthermore, we select a mini-batch of 64 hard positive and 64 hard negative cell pairs for the cell merging module.  

In the testing phase of table detection, the shorter side of each testing image is rescaled to be 512 pixels with the longer side not exceeding 1,024 pixels. {\color{black}{We set the number of selected corners (top-$K$) as 100 with a corner score threshold $C_{th}$ as 0.3.}} We apply the standard NMS algorithm with an IoU threshold of 0.3 on the detected tables to suppress redundant detections. For TSR testing, we rescale the longer side of each cropped table image to be 1,024 pixels while keeping the aspect ratio, except for the SciTSR dataset where the cropped images are not resized. The binarization score threshold $S_{th}$ is set as 0.8. The grid cells from the split model are merged based on the merging scores with a threshold of 0.8. {\color{black}{Abovedmentioned hyper-parameters are tuned on our in-house dataset, and we directly apply them to other datasets without further tuning.}}

\subsection{Experiments on table detection}
\label{subsec:exp_TD}
\subsubsection{Comparisons with prior arts}
\begin{figure}[t]
    \centering
    \setlength{\abovecaptionskip}{-0.2cm}
    \includegraphics[width=1.0\linewidth]{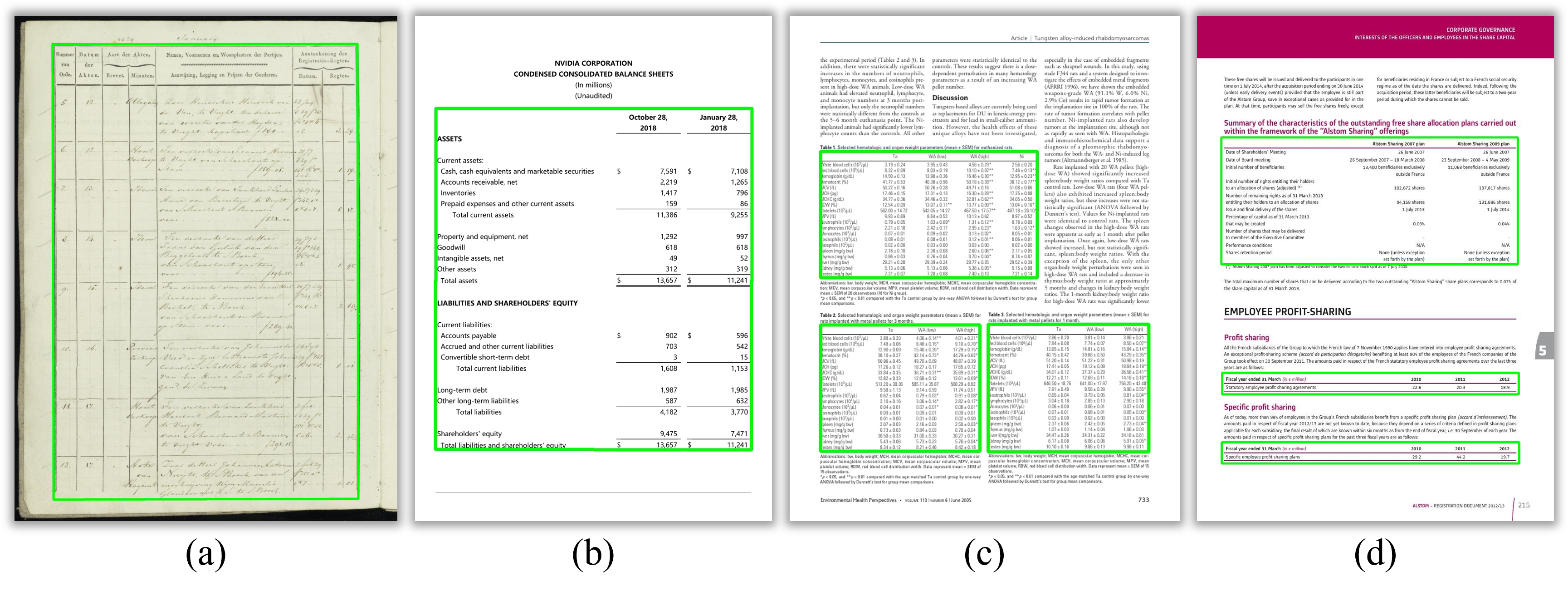}
    \caption{Qualitative results of our table detector. (a-b) are from cTDaR TrackA, (c) is from PubLayNet, and (d) is from IIIT-AR-13K.}
    \label{fig:table_detection_demo}
\end{figure}

We compare our table detection approach with other most competitive methods on cTDaR TrackA, PubLayNet and IIIT-AR-13K. All the results of our approach are based on single-model and single-scale testing. The results are listed in Table~\ref{tab:cTDaR_TrackA}, Table~\ref{tab:PubLayNet} and Table~\ref{tab:IIIT-AR-13K}. On cTDaR TrackA, our approach achieves the best WAvg. F1-score of 94.9\%, outperforming other methods by a notable margin. Furthermore, it is noted that our approach has achieved the best F1-scores at higher IoU thresholds, {\color{black}{e.g., 92.9\% vs. 91.5\% with the IoU threshold at 0.9}}, which demonstrates the superiority of our approach on high precision table localization. On PubLayNet, our model with the ResNet-18 backbone network can even substantially outperform the Mask R-CNN model with the ResNeXt-101 backbone network {\color{black}{by improving the $AP^{0.5:0.95}$ from 96.0\% to 97.0\%, and significantly improving the $AP^{0.95}$ from 81.4\% to 92.0\%.}} Similarly, on IIIT-AR-13K, our model can also substantially outperform the Mask R-CNN model with the ResNet-101 backbone network by improving the AP from 97.6\% to 98.2\% on the validation set and from 96.5\% to 97.7\% on the testing set, respectively. To push the table detection performance of the Cascade Mask R-CNN framework on public benchmarks, Agarwal et al. \cite{agarwal2021cdec} employed a more powerful backbone network, i.e., dual backbone ResNeXt-101 with deformable convolution filters. However, its performance is still inferior to ours on cTDaR TrackA and PubLayNet. The superior performance achieved on these public benchmark datasets shows the effectiveness and robustness of our approach. Some qualitative results of our approach on these datasets are presented in Fig.~\ref{fig:table_detection_demo}. 

\subsubsection{Ablation study}
\label{sebsebsec:abl_corner}
\begin{figure}[t]
    \centering
    \setlength{\abovecaptionskip}{-0.2cm}
    \includegraphics[width=1.0\linewidth]{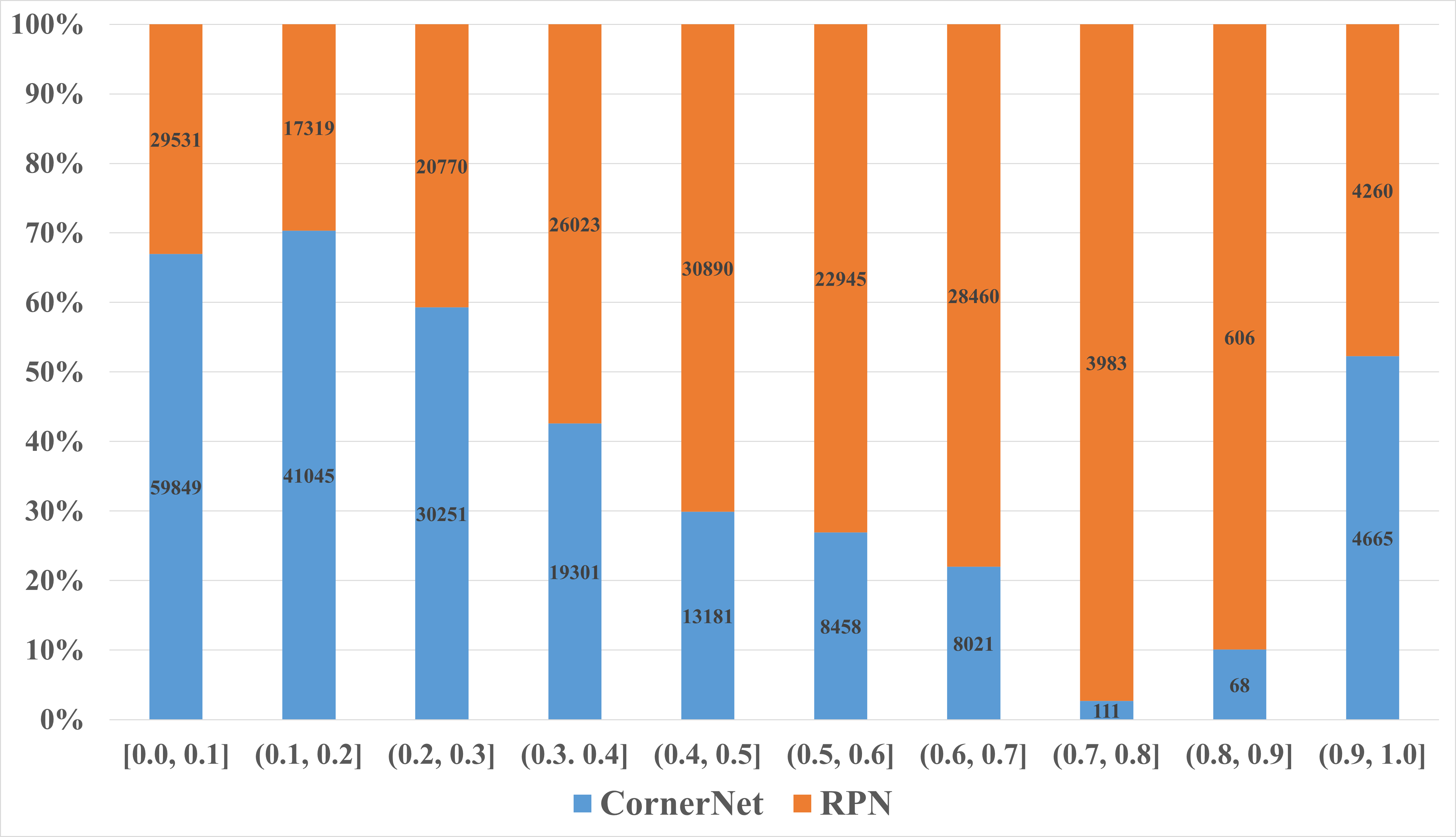}
    \caption{{\color{black}{The distribution of the IoU between proposals and GT boxes, where the x-axis represents the ranges of IoU, and the y-axis represents the ratio of the number of proposals in the corresponding IoU range between the two methods.}}}
    \label{fig:proposal_analysis}
\end{figure}

{\color{black}{\textbf{CornerNet vs. RPN for table proposal generation.} To compare the proposed CornerNet based table proposal generation algorithm with RPN \cite{ren2015faster}, we evaluate their recall rates with top-50 proposals on PubLayNet first. The quantitative results are given in Table~\ref{tab:proposal_analysis}, which shows that although these two methods can achieve a similar recall rate at the IoU threshold of 0.7, the CornerNet based method can significantly outperform RPN under a higher IoU threshold 0.9, i.e., 97.8\% vs. 89.3\%. Then, we further evaluate the end-to-end performance and the comparison results are given in the last two rows of Table~\ref{tab:PubLayNet}. We can find that the performance of RPN based table detector is inferior to our CornerNet based detector, which shows that the quality of the proposals generated by CornerNet is better than RPN. To reveal the relation between proposal quality and end-to-end detection accuracy, we further compute the maximum IoU between each proposal and all the GT boxes, and the corresponding statistical results are shown in Fig.~\ref{fig:proposal_analysis}. We find that there are more proposals from RPN within the IoU range of (0.7, 0.9]. As the proposals in this range will also be taken as positive samples during the training of Fast R-CNN, these low quality proposals will also have high classification scores and survive from the NMS step, which will degrade the end-to-end performance when evaluating at high IoU thresholds. Compared with RPN, the percentage of well-localized proposals (IoU$>$0.9) in the positive samples (IoU$>$0.7) from CornerNet is much higher (96.3\% vs. 48.1\%), which contributes to better end-to-end table detection performance. These experimental results demonstrate the superiority of our CornerNet based table proposal generation algorithm for achieving higher localization accuracy and better end-to-end table detection results.
}}

\begin{table}[t]
    \centering
    \footnotesize
    \color{black}{
    \caption{Table proposal generation quality comparison on PubLayNet.}
    \label{tab:proposal_analysis}
    \begin{tabular}{ c  c  c  c  c }
        \toprule
        \multirow{2}{*}{Methods} & {IoU@0.6} & {IoU@0.7} & {IoU@0.8} & {IoU@0.9}\\
        & {Recall(\%)} & {Recall(\%)} & {Recall(\%)} & {Recall(\%)}\\
        \midrule
        RPN & 99.5 & 99.3 & 98.8 & 89.3\\
        CornerNet & 99.5 & 99.4 & 99.2 & \textbf{97.8}\\
        \bottomrule
    \end{tabular}
    }
\end{table}

\subsection{Experiments on table structure recognition}
\begin{table}[t]
    \setlength{\tabcolsep}{4.5pt}
    \footnotesize
    \centering
    \caption{TSR performance comparison on SciTSR and SciTSR-COMP. * indicates that the results are from\cite{chi2019complicated}.}
    \label{tab:SciTSR}
    {\color{black}{
    \begin{tabular}{ c  c  c  c  c  c  c  c }
        \hline
        \multirow{2}{*}{Methods} & \multicolumn{3}{c}{SciTSR(\%)} && \multicolumn{3}{c}{SciTSR-COMP(\%)} \\\cline{2-4}\cline{6-8}
         & P & R & F1 && P & R & F1 \\
        \hline
        Adobe$^*$ & 93.0 & 78.4 & 85.1 && 90.1 & 71.7 & 79.8 \\
        DeepDeSRT\cite{schreiber2017deepdesrt}$^*$ & 90.6 & 88.7 & 89.0 && 86.3 & 83.1 & 84.6 \\
        Tabby\cite{shigarov2016configurable}$^*$ & 92.6 & 92.0 & 92.1 && 89.2 & 87.2 & 88.2 \\
        TabStruct-Net\cite{raja2020table} & 92.7 & 91.3 & 92.0 && 90.9 & 88.2 & 89.5 \\
        GraphTSR\cite{chi2019complicated} & 95.9 & 94.8 & 95.3 && 96.4 & 94.5 & 95.5 \\
        SEM\cite{zhang2022split} & 97.7 & 96.5 & 97.1 && 96.8 & 94.7 & 95.7 \\
        LGPMA\cite{qiao2021lgpma} & 98.2 & \textbf{99.3} & 98.8 && 97.3 & \textbf{98.7} & 98.0 \\
        \hline
        \textbf{Ours} & \textbf{99.4} & 99.1 & \textbf{99.3} && \textbf{99.0} & 98.4 & \textbf{98.7} \\
        \hline
    \end{tabular}
    }}
\end{table}
\begin{table}[t]
    \footnotesize
    \centering
    \caption{TSR performance comparison on the validation set of PubTabNet.} 
    \label{tab:PubTabNet}
    {\color{black}{
    \begin{tabular}{ c  c  c }
        \hline
        Methods & TEDS(\%) & TEDS-Struct(\%) \\
        \hline
        EDD\cite{zhong2020image} & 88.3 & - \\
        TabStruct-Net\cite{raja2020table} & - & 90.1 \\
        GTE\cite{zheng2021global} & - & 93.0 \\
        LGPMA\cite{qiao2021lgpma} & 94.6 & 96.7 \\
        \hline
        \textbf{Ours} & - & \textbf{97.0} \\
        \hline
    \end{tabular}
    }}
\end{table}

\begin{table*}[h!]
    \setlength{\tabcolsep}{4.5pt}
    \footnotesize
    \centering
    \color{black}{
    \caption{TSR Performance comparison on ICDAR2019 cTDaR TrackB2-Modern. * indicates that the results are from \cite{gao2019icdar}.}
    \label{tab:cTDaR_TrackB2_Official_100}
    \begin{tabular}{ c  c  c  c  c  c  c  c  c  c  c  c  c  c  c  c  c}
        \toprule
        \multirow{2}{*}{Methods} & \multicolumn{3}{c}{IoU@0.6(\%)} && \multicolumn{3}{c}{IoU@0.7(\%)} &&
        \multicolumn{3}{c}{IoU@0.8(\%)} &&
        \multicolumn{3}{c}{IoU@0.9(\%)} & WAvg. \\\cline{2-4}\cline{6-8}\cline{10-12}\cline{14-16}
         & P & R & F1 && P & R & F1 && P & R & F1 && P & R & F1 & F1(\%)\\
        \midrule
        Zou et al.\cite{zou2020deep} & 18.8 & 10.1 & 13.1 && - & - & - && 1.7 & 0.9 & 1.2 && - & - & - & - \\
        NLPR-PAL* & 32.2 & 42.1 & 36.5 && 26.9 & 35.1 & 30.5 && 17.2 & 22.5 & 19.5 && 3.1 & 4.0 & 3.5 & 20.6 \\
        CascadeTabNet\cite{prasad2020cascadetabnet} & 49.9 & 39.0 & 43.8 && 40.3 & 31.5 & 35.4 && 21.6 & 16.9 & 19.0 && 4.1 & 3.2 & 3.6 & 23.2 \\
        GTE\cite{zheng2021global} & - & - & 38.5 && - & - & - && - & - & - && - & - & - & 24.8 \\
        \midrule
        \textbf{Ours} & \textbf{76.4} & \textbf{76.8} & \textbf{76.6} && \textbf{71.3} & \textbf{71.6} & \textbf{71.4} && \textbf{58.1} & \textbf{58.4} & \textbf{58.3} && \textbf{25.7} & \textbf{25.8} & \textbf{25.8} & \textbf{55.3} \\
        \bottomrule
    \end{tabular}
    }
\end{table*}
\subsubsection{Comparisons with prior arts}
\begin{figure}[t]
    \centering
    \setlength{\abovecaptionskip}{-0.5cm}
    \includegraphics[width=1.0\linewidth]{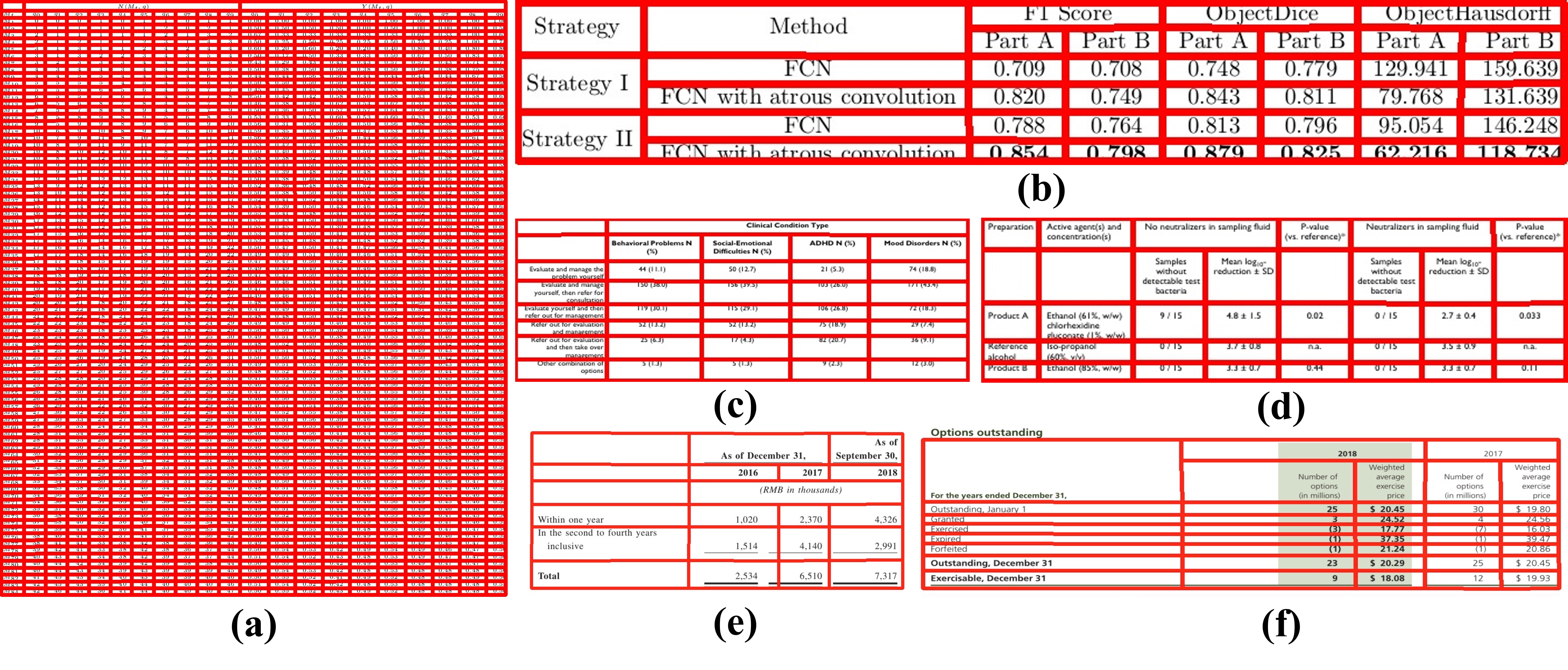}
    \caption{Qualitative results of our table structure recognition approach. (a-b) are from SciTSR, (c-d) are from PubTabNet, {\color{black}{(e-f) are cropped from cTDaR TrackB2-Modern.}}}
    \label{fig:TSR_demo}
\end{figure}

We compare our table structure recognition approach with other most competitive methods on SciTSR, PubTabNet and {\color{black}{cTDaR TrackB2-Modern}}. 
On SciTSR and SciTSR-COMP (see Table~\ref{tab:SciTSR}), our approach has achieved state-of-the-art performance {\color{black}{with the best F1-score of 99.3\% and 98.7\% on the full testing set and the complicated subset, respectively.}} Moreover, our approach shows negligible performance degradation on the complicated subset, which demonstrates its robustness to tables with complex structures. Similarly, on PubTabNet (see Table~\ref{tab:PubTabNet}), {\color{black}{our approach has also achieved the best TEDS-Struct score of 97.0\%.}} It is noted that, the recent best performing method LGPMA \cite{qiao2021lgpma} (the winner of ICDAR 2021 Competition on Scientific Literature Parsing Task B \cite{jimeno2021icdar}) has leveraged an important task constraint, namely tables are axis-aligned, to achieve higher accuracy. So, it cannot be directly applied to distorted tables. Our approach doesn’t rely on such kind of assumptions but still achieves higher accuracy. {\color{black}{On cTDaR TrackB2-Modern, our table detector and table structure recognizer are combined together to conduct end-to-end evaluation. Since the outputs of our approach are cell boxes rather than convex hulls of cell contents, for the sake of fair comparison, we use the same text detection algorithm as CascadeTabNet \cite{prasad2020cascadetabnet} to detect texts in each image and then assign them to table cells if 80\% of a text box is located in a cell box. 
As shown in Table~\ref{tab:cTDaR_TrackB2_Official_100}, our approach surpasses previous methods by a large margin.
}}
Some qualitative results of our approach on these datasets are presented in Fig.~\ref{fig:TSR_demo}.

\begin{figure}[h!]
    \centering
    \setlength{\abovecaptionskip}{-0.3cm}
    \includegraphics[width=1.0\linewidth]{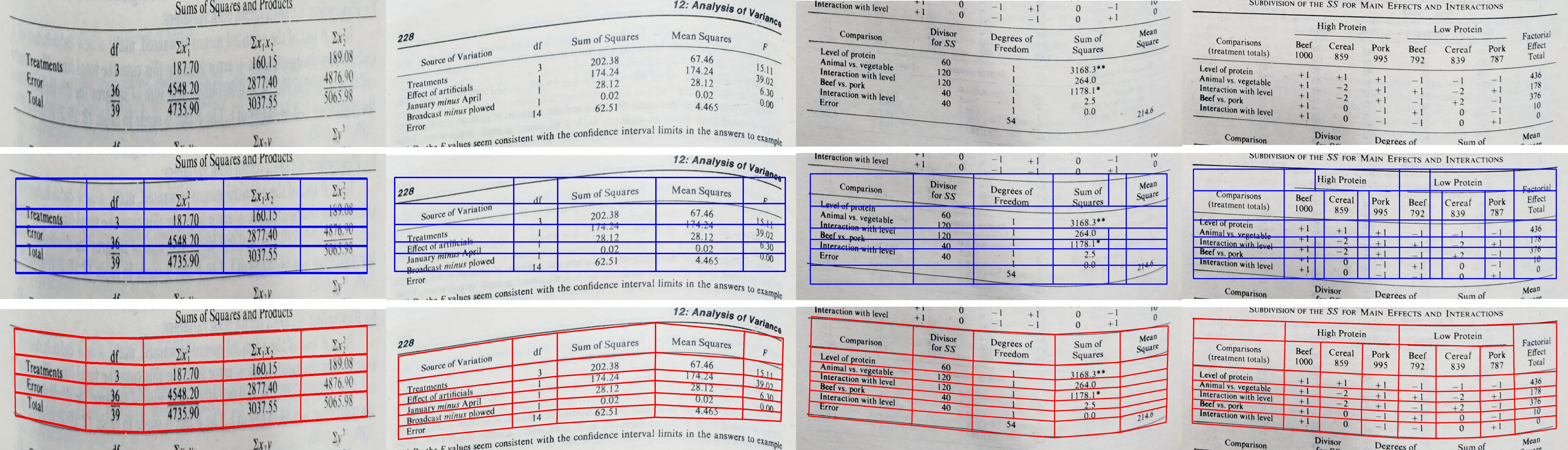}
    \caption{Qualitative results on the in-house dataset. $1^{st}$ row: original images; $2^{nd}$ row: results from SPLERGE \cite{tensmeyer2019deep}; $3^{rd}$ row: results from our table structure recognizer.}
    \label{fig:TSR_curved_demo}
\end{figure}

To further validate the robustness of our approach to distorted or even curved table images, we conducted experiments on the in-house dataset and compared our table structure recognizer with SPLERGE. As shown in Table~\ref{tab:private_dataset}, our approach outperforms SPLERGE significantly by improving the WAvg. F1-score from 63.8\% to 94.6\%. Some qualitative results of our approach on this challenging dataset are presented in Fig.~\ref{fig:TSR_curved_demo}, from which we can observe that our table structure recognizer can work robustly under various challenging conditions such as tables without ruling lines, tables with empty or spanning cells and distorted or even curved shapes. 
\begin{table*}[t]
    \setlength{\tabcolsep}{4.5pt}
    \footnotesize
    \centering
    \caption{TSR performance comparison on the in-house dataset.}
    \label{tab:private_dataset}
    \begin{tabular}{ c  c  c  c  c  c  c  c  c  c  c  c  c  c  c  c  c}
        \toprule
        \multirow{2}{*}{Methods} & \multicolumn{3}{c}{IoU@0.6(\%)} && \multicolumn{3}{c}{IoU@0.7(\%)} &&
        \multicolumn{3}{c}{IoU@0.8(\%)} &&
        \multicolumn{3}{c}{IoU@0.9(\%)} & WAvg. \\\cline{2-4}\cline{6-8}\cline{10-12}\cline{14-16}
         & P & R & F1 && P & R & F1 && P & R & F1 && P & R & F1 & F1(\%)\\
        \midrule
        SPLERGE \cite{tensmeyer2019deep} & 75.9 & 55.2 & 63.9 && 75.8 & 55.1 & 63.8 && 75.7 & 55.0 & 63.7 && 75.7 & 55.0 & 63.7 & 63.8 \\
        \textbf{Ours} & \textbf{94.9} & \textbf{94.5} & \textbf{94.7} && \textbf{94.8} & \textbf{94.4} & \textbf{94.6} && \textbf{94.8} & \textbf{94.4} & \textbf{94.6} && \textbf{94.7} & \textbf{94.3} & \textbf{94.5} & \textbf{94.6} \\
        \bottomrule
    \end{tabular}
\end{table*}
\begin{table}[t]
    \setlength{\tabcolsep}{4.5pt}
    \centering
    \footnotesize
    \caption{Ablation study of the kernel width for spatial CNN modules.}
    \label{tab:kernel_size}
    \begin{tabular}{c c c c c c c}
        \hline
        Kernel Width ($k$) & 1 & 3 & 5 & 7 & 9 & 11  \\
        \hline
        WAvg. F1-score (\%) & 93.9 & 94.8 & 95.3 & 95.8 & \textbf{96.0} & 95.7 \\
        \hline
    \end{tabular}
\end{table}

\subsubsection{Ablation study}
\textbf{Influence of kernel width in spatial CNN modules.} We investigate the influence of the kernel width in the four spatial CNN modules to TSR accuracies. The experimental results on the manually labeled cTDaR modern subset are shown in Table~\ref{tab:kernel_size}. Here, the kernel width determines the number of pixels from which a pixel could receive messages directly. Consistent with the observations in \cite{pan2018spatial}, increasing the kernel width improves the performance up to a saturation point ($k=9$), and then the performance slightly decreases. Therefore, we set the kernel width as 9 for all the other experiments.

\begin{figure}[h!]
    \centering
    \setlength{\abovecaptionskip}{-0.2cm}
    \includegraphics[width=1.0\linewidth]{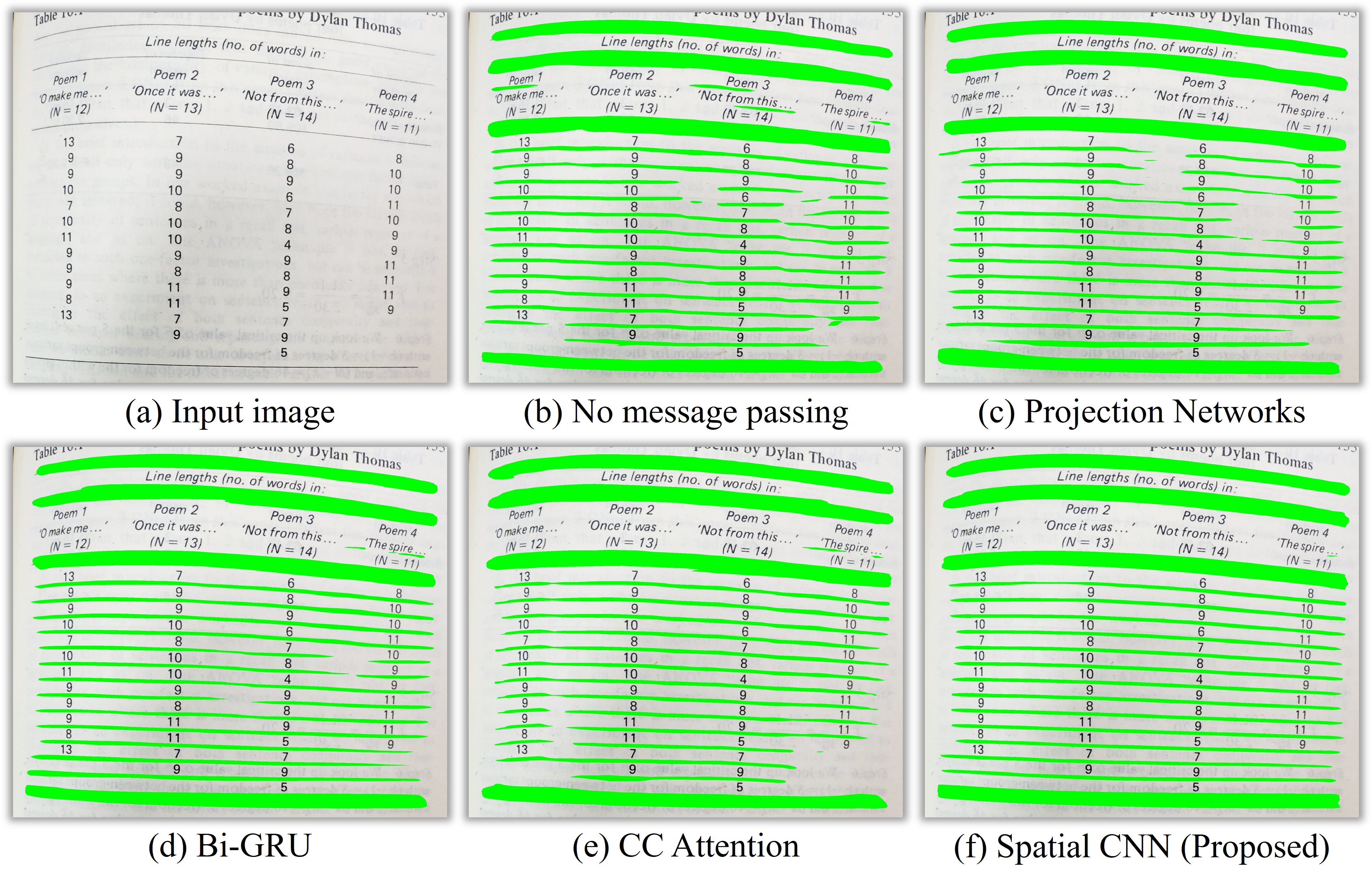}
    \caption{Some comparison examples from different message passing methods for separation line prediction.}
    \label{fig:scnn_demo}
\end{figure}

\textbf{Effectiveness of spatial CNN based separation line prediction.} We compare our spatial CNN based message passing method with two previously used methods in the TSR field, i.e., projection networks \cite{tensmeyer2019deep} and Bi-GRU \cite{khan2019table}. Moreover, as self-attention operations are also known to be good at aggregating global context information, we also select a representative one, i.e., criss-cross attention \cite{huang2019ccnet}, for comparison. All models are trained with the same hyper-parameters for fair comparison and tested on our challenging in-house dataset. We have also implemented a baseline model, i.e., removing the spatial CNN modules directly from our TSR model. The quantitative results of these variants are given in Table~\ref{tab:scnn_ablation_study} and some qualitative results are shown in Fig.~\ref{fig:scnn_demo}. The experimental results show that the performance of other message passing methods are obviously inferior to our spatial CNN based method, especially for tables with large blank spaces or curved tables, which can demonstrate the effectiveness of our spatial CNN based separation line prediction method. 
\begin{table*}[t]
    \setlength{\tabcolsep}{4.5pt}
    \footnotesize
    \centering
    \caption{Comparison of different message passing methods.}
    \label{tab:scnn_ablation_study}
    \begin{tabular}{ c  c  c  c  c  c  c  c  c  c  c  c  c  c  c  c  c}
        \toprule
        Message Passing & \multicolumn{3}{c}{IoU@0.6(\%)} && \multicolumn{3}{c}{IoU@0.7(\%)} &&
        \multicolumn{3}{c}{IoU@0.8(\%)} &&
        \multicolumn{3}{c}{IoU@0.9(\%)} & WAvg. \\\cline{2-4}\cline{6-8}\cline{10-12}\cline{14-16}
        Methods & P & R & F1 && P & R & F1 && P & R & F1 && P & R & F1 & F1(\%)\\
        \midrule
        No message passing & 92.9 & 92.2 & 92.5 && 92.8 & 92.1 & 92.4 && 92.7 & 92.0 & 92.4 && 92.7 & 91.9 & 92.3 & 92.4 \\
        Projection Networks \cite{tensmeyer2019deep} & 93.8 & 92.6 & 93.2 && 93.7 & 92.5 & 93.1 && 93.6 & 92.5 & 93.0 && 93.5 & 92.4 & 93.0 & 93.0 \\
        Bi-GRU (2 layers) \cite{khan2019table} & 93.7 & 92.7 & 93.2 && 93.7 & 92.6 & 93.1 && 93.6 & 92.5 & 93.1 && 93.6 & 92.5 & 93.0 & 93.1 \\
        CC Attention \cite{huang2019ccnet} & 94.1 & 93.7 & 93.9 && 94.1 & 93.6 & 93.8 && 94.0 & 93.5 & 93.8 && 94.0 & 93.5 & 93.7 & 93.8 \\
        \textbf{Spatial CNN (Proposed)} & \textbf{94.9} & \textbf{94.5} & \textbf{94.7} && \textbf{94.8} & \textbf{94.4} & \textbf{94.6} && \textbf{94.8} & \textbf{94.4} & \textbf{94.6} && \textbf{94.7} & \textbf{94.3} & \textbf{94.5} & \textbf{94.6} \\
        \bottomrule
    \end{tabular}
\end{table*}
\begin{table*}[t]
    \setlength{\tabcolsep}{4.5pt}
    \footnotesize
    \centering
    \caption{Comparison of different cell merging methods.}
    \label{tab:grid_cnn_ablation_study}
    \begin{tabular}{ c  c  c  c  c  c  c  c  c  c  c  c  c  c  c  c  c}
        \toprule
        Cell Merging & \multicolumn{3}{c}{IoU@0.6(\%)} && \multicolumn{3}{c}{IoU@0.7(\%)} &&
        \multicolumn{3}{c}{IoU@0.8(\%)} &&
        \multicolumn{3}{c}{IoU@0.9(\%)} & WAvg. \\\cline{2-4}\cline{6-8}\cline{10-12}\cline{14-16}
        Methods & P & R & F1 && P & R & F1 && P & R & F1 && P & R & F1 & F1(\%)\\
        \midrule
        No cell merging & 91.7 & 90.5 & 91.1 && 91.6 & 90.4 & 91.0 && 91.5 & 90.3 & 90.9 && 91.5 & 90.2 & 90.9 & 90.9 \\
        Relation Network \cite{ma2021relatext} & 93.5 & 93.1 & 93.3 && 93.4 & 93.0 & 93.2 && 93.3 & 93.0 & 93.1 && 93.3 & 92.9 & 93.1 & 93.2 \\
        GCN \cite{ma2021relatext} & 94.2 & 94.0 & 94.1 && 94.1 & 93.9 & 94.0 && 94.1 & 93.8 & 94.0 && 94.0 & 93.8 & 93.9 & 94.0 \\
        \textbf{Grid CNN (Proposed)} & \textbf{94.9} & \textbf{94.5} & \textbf{94.7} && \textbf{94.8} & \textbf{94.4} & \textbf{94.6} && \textbf{94.8} & \textbf{94.4} & \textbf{94.6} && \textbf{94.7} & \textbf{94.3} & \textbf{94.5} & \textbf{94.6} \\
        \bottomrule
    \end{tabular}
\end{table*}

\textbf{Effectiveness of Grid CNN based cell merging.} We further compare the proposed Grid CNN based cell merging method with other two visual relationship prediction based methods \cite{ma2021relatext}, which are based on relation network and GCN respectively, on the in-house dataset to demonstrate the effectiveness of Grid CNN for cell merging. The experimental results are listed in Table~\ref{tab:grid_cnn_ablation_study}, from which we can find that the cell merging module can significantly improve the performance of our TSR model (90.9\% vs. 94.6\%). Moreover, the last three rows show that the proposed Grid CNN based cell merging method is more effective than the relation network based (93.2\% vs. 94.6\%) and the GCN based (94.0\% vs. 94.6\%) methods. Based on our observations, due to the grid arrangement of cell features, Grid CNN can leverage context information effectively with several stacked convolution layers to improve cell merging accuracy, leading to improved robustness to tables with hierarchical spanning cells.

\subsection{Limitations of our approach}
\label{subsec:limitations}
\begin{figure}[t]
    \centering
    \setlength{\abovecaptionskip}{-0.2cm}
    \includegraphics[width=1.0\linewidth]{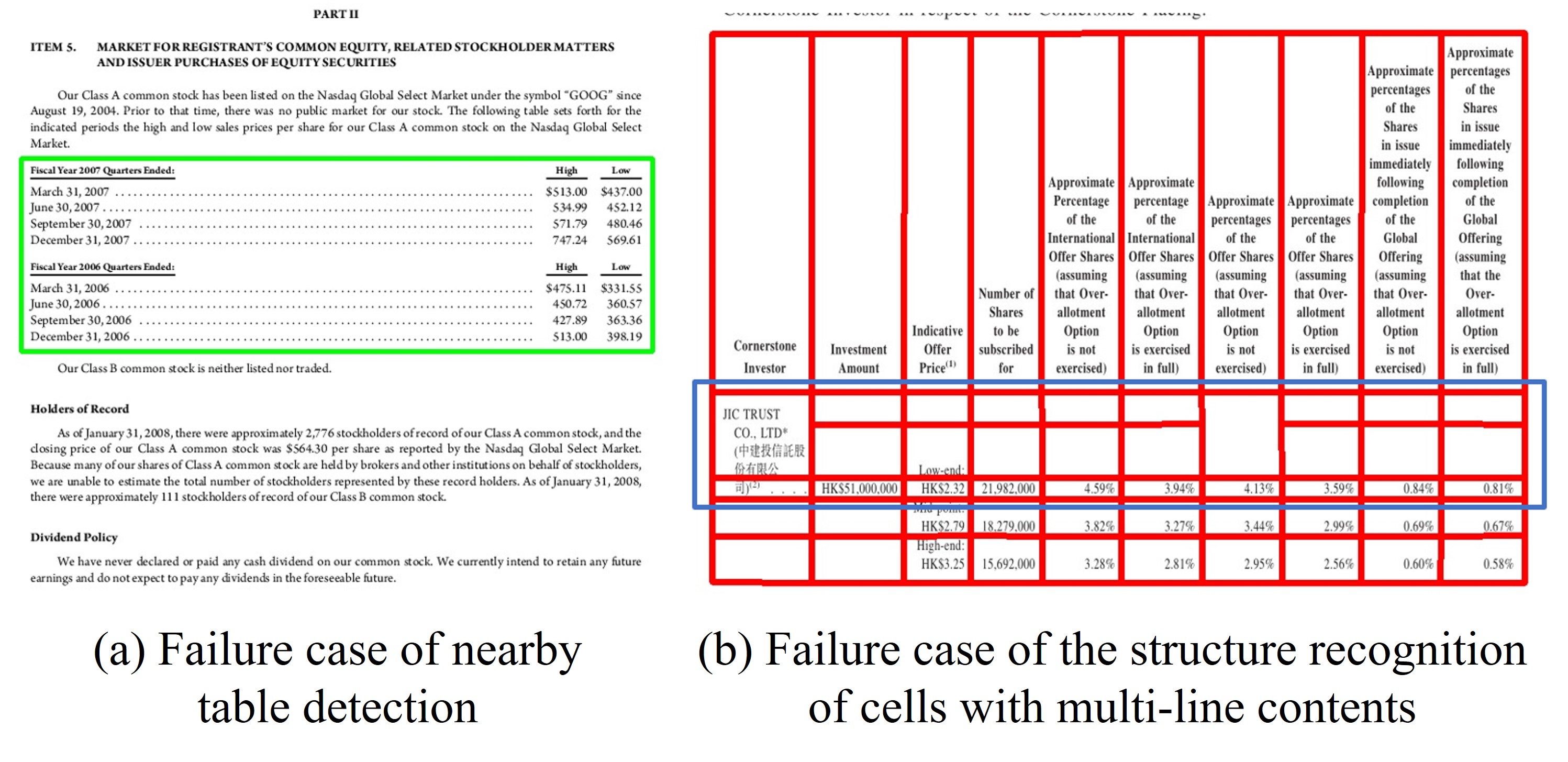}
    \caption{Some typical failure cases, including the detection of nearby tables and the structure recognition of cells with multi-line contents. }
    \label{fig:failure_cases}
\end{figure}

Although the proposed RobusTabNet shows superior capability in most scenarios as demonstrated in the previous experiments, it still has some limitations. For example, our current table detector still struggles with nearby tables, and our table structure recognizer is not robust enough to cells with multi-line contents. Some failure examples are presented in Fig.~\ref{fig:failure_cases}. {\color{black}{Furthermore, our TSR approach will fail on some extremely dense tables, because the predicted segmentation 
masks of nearby separation lines could be overlapped.}} Note that these difficulties are common challenges for other state-of-the-art methods. Finding effective solutions to these problems will be our future work. {\color{black}{Moreover, since the tables in existing datasets are mostly with black lines/letters and white backgrounds, the effectiveness and generalization ability of our approach on tables with different types of backgrounds, text fonts and line colors need to be studied in the future.}}

\section{Conclusion and future work}
\label{sec:conclusion}
In this paper, we introduce a new table detection and structure recognition approach named RobusTabNet to extract tables from heterogeneous document images. For table detection, we use CornerNet as a new region proposal network for Faster R-CNN, which can leverage more precise corner points generated from heatmaps to improve table localization accuracy. For table structure recognition, we propose two effective techniques to significantly improve the capability of the split-and-merge paradigm, i.e., spatial CNN based separation line prediction and Grid CNN based cell merging. As the spatial CNN can effectively propagate contextual information across the whole table image, improved robustness can be achieved to tables with large blank spaces and curved tables. Moreover, as the whole table is compactly represented as a grid, a simple but effective Grid CNN can be used to achieve excellent cell merging accuracy. Consequently, the proposed RobusTabNet has achieved state-of-the-art performance on both table detection (cTDaR TrackA, PubLayNet and IIIT-AR-13K) and structure recognition (SciTSR, PubTabNet and {\color{black}{cTDaR TrackB2-Modern}}) public benchmarks. We have further validated the robustness of our approach to tables with complex structures, large blank spaces, as well as distorted or even curved shapes on a more challenging in-house dataset.

For future work, we will study how to leverage header analysis techniques to disambiguate nearby tables. Furthermore, we will also explore how to incorporate textual information into our Grid CNN module to improve the robustness of our table structure recognizer to cells with multi-line contents. {\color{black}{To achieve more robust structure recognition of dense tables, we will study effective technologies for adaptive scaling.}} As for latency reduction, we will explore an end-to-end solution for table extraction, where the table detector and the table structure recognizer can share a same backbone network.

\bibliography{main.bbl}
\end{document}